\documentclass{article}

\usepackage{microtype}
\usepackage{graphicx}
\usepackage{subfigure}
\usepackage{booktabs} 
\usepackage{xspace}
\usepackage{multirow}
\usepackage{todonotes}
\usepackage{amsmath}
\usepackage{amssymb}
\usepackage{amsfonts}
\usepackage{bm}
\usepackage{appendix}
\usepackage[hyphens]{url}
\usepackage{wrapfig}

\usepackage{hyperref}

\usepackage[accepted]{icml2020}

\icmltitlerunning{ACFlow: Flow Models for Arbitrary Conditional Likelihoods}

\newcommand{\method}{ACFlow\xspace}
\newcommand{\tabpm}{$\pm$}

\begin{document}

\twocolumn[
\icmltitle{ACFlow: Flow Models for Arbitrary Conditional Likelihoods}



\icmlsetsymbol{equal}{*}

\begin{icmlauthorlist}
\icmlauthor{Yang Li}{unc}
\icmlauthor{Shoaib Akbar}{ncsu}
\icmlauthor{Junier B. Oliva}{unc}
\end{icmlauthorlist}

\icmlaffiliation{unc}{Department of Computer Science, University of North Carolina at Chapel Hill, NC, USA}
\icmlaffiliation{ncsu}{Department of Mathematics, North Carolina State University, NC, USA}

\icmlcorrespondingauthor{Yang Li}{yangli95@cs.unc.edu}

\icmlkeywords{Generative Models, Normalizing Flows, Marginalization}

\vskip 0.3in
]



\printAffiliationsAndNotice{}  

\begin{abstract}
Understanding the dependencies among features of a dataset is at the core of most unsupervised learning tasks. However, a majority of generative modeling approaches are focused solely on the joint distribution $p(x)$ and utilize models where it is intractable to obtain the conditional distribution of some arbitrary subset of features $x_u$ given the rest of the observed covariates $x_o$: $p(x_u \mid x_o)$. 
Traditional conditional approaches provide a model for a \emph{fixed} set of covariates conditioned on another \emph{fixed} set of observed covariates.
Instead, in this work we develop a model that is capable of yielding \emph{all} conditional distributions $p(x_u \mid x_o)$ (for arbitrary $x_u$) via tractable conditional likelihoods.
We propose a novel extension of (change of variables based) flow generative models, arbitrary conditioning flow models (\method). \method can be conditioned on arbitrary subsets of observed covariates, which was previously infeasible.
We further extend \method to model the joint distributions $p(x)$ and arbitrary marginal distributions $p(x_u)$. 
We also apply \method to the imputation of features, and develop a unified platform for both multiple and single imputation by introducing an auxiliary objective that provides a principled single ``best guess'' for flow models.
Extensive empirical evaluations show that our model achieves state-of-the-art performance in modeling arbitrary conditional distributions in addition to both single and multiple imputation in synthetic and real-world datasets.
\end{abstract}

\section{Introduction}
Spurred on by recent impressive results, there has been a surge in interest for generative probabilistic modeling in machine learning.
These models learn an approximation of the underlying data distribution and are capable of drawing realistic samples from it. 
Generative models have a multitude of potential applications, including image restoration \cite{ledig2017photo}, agent planning \cite{houthooft2016vime}, and unsupervised representation learning \cite{chen2016infogan}.

Most generative approaches are solely focused on the joint distribution of features, $p(x)$, and are opaque in the conditional dependencies that are carried among subsets of features. 
Existing conditional generative models are mostly conditioned on a fixed set of covariates, such as class labels \cite{kingma2018glow} or other data points \cite{li2019forest}.
In this work, we propose a framework, arbitrary conditioning flow models (\method), to construct generative models that 
yield tractable (analytically available) conditional likelihoods $p( x_u \mid x_o)$ of an arbitrary subset of covariates, $x_u$, given the remaining observed covariates $x_o$. Although the complete data $x$ comes from a certain distribution, all its conditional distributions $p(x_u \mid x_o)$ vary when conditioned on different $x_o$, which poses challenges for modeling the highly multimodal distributions. Furthermore, the dimensionality of $x_u$ and $x_o$ could be arbitrary. 

Dealing with arbitrary dimensionality is a largely unexplored topic in current generative models. One might want to explicitly learn a separate model for each different subset of observed covariates; however, this approach quickly becomes infeasible as it requires an exponential number of models with respect to the dimensionality of the input space.
In this work, we propose several conditional transformations that handle arbitrary dimensionality in a principled manner and further combine them with an autoregressive likelihood approach for flexible, tractable generative modeling.

In addition, \method can handle the joint distribution $p(x)$ and arbitrary marginal distributions $p(x_u)$ as special cases. Joint distribution $p(x)$ can be obtained by conditioning on an empty set, i.e., $p(x \mid \emptyset)$, while arbitrary marginal distribution $p(x_u)$ can be obtained similarly as $p(x_u \mid \emptyset)$. In effect, this allows for our model to perform arbitrary marginalization, which has previously been infeasible in flow models and autoregressive frameworks.

Besides computing likelihoods, we also explore the use of \method for imputation, where we infer possible values of $x_u$, given observed values $x_o$ both in general real-valued data and images (for inpainting). From the perspective of probabilistic modeling, data imputation attempts to learn a distribution of the unobserved covariates, $x_u$, given the observed covariates, $x_o$.
Thus, generative modeling is a natural fit for data imputation. It handles single imputation and multiple imputation in an unified framework by allowing the generation of an arbitrary number of samples. More importantly, it quantifies the uncertainty in a principled manner.

Our contributions are as follows. 1) We propose a novel extension of flow-based generative models to model the conditional distribution of arbitrary unobserved covariates. Our method is the first to develop invertible transformations that operate on an arbitrary set of covariates. 2) We strengthen a flow-based model by using a novel autoregressive conditional likelihood. 3) We propose a novel penalty to generate a single imputed ``best guess" for models without an analytically available mean. 4) We extend ACFlow to model arbitrary marginals, enabling one to do approximate marginalization of flow models, which was previously infeasible. 5) We run extensive empirical studies and show that \method achieves state-of-the-art arbitrary conditional likelihoods on benchmark datasets.

\section{Problem Formulation}
Consider a real-valued\footnote{Data with categorical dimensions can be handled by a special case of our model, please refer to appendix~\ref{sec:categorical}.} distribution $p(x)$ over $\mathbb{R}^d$.
We are interested in estimating the conditional distribution of \emph{all} possible subsets of covariates $u \subseteq \{1, \ldots, d\}$ conditioned on the remaining observed covariates $o = \{1, \ldots, d\} \setminus u$. That is, we shall estimate $p(x_u \mid x_o)$ where $x_u \in \mathbb{R}^{|u|}$ and $x_o \in \mathbb{R}^{|o|}$, for all possible subsets $u$.

For ease of notation, let $b \in \{0, 1\}^d$ be a binary mask  indicating which dimensions are observed.
Furthermore, let $v[b]$ index a vector $v$ using a bitmask $b$. Thus, $x_o=x\left[b\right]$ denotes observed dimensions and $x_u = x\left[1-b\right]$ denotes unobserved dimensions. 
We also apply this indexing mechanism to matrices such that $W[b, b]$ indexes rows and then columns.
Without loss of generality, conditionals may also be conditioned on the bitmask $b$, $p(x_u \mid x_o, b)$, and will be estimated with maximum log-likelihood estimation as described below. 
In addition, imputation tasks shall be accomplished by generating samples from the conditional distributions $p(x_u \mid x_o, b)$.

\section{Background}

\method builds on Transformation Autoregressive Networks (TANs) \cite{oliva2018transformation}, a flow-based model that combines transformation of variables with autoregressive likelihoods.
We expound on flow-based models and TANs below.

The change of variable theorem \eqref{eq:changevariables} is the cornerstone of flow-based generative models, where $q$ represents an invertible transformation that transforms covariates from input space $\mathcal{X}$ into a latent space $\mathcal{Z}$.
\begin{equation}\label{eq:changevariables}
    p_{\mathcal{X}}(x) = \left| \det \frac{dq}{dx} \right| p_{\mathcal{Z}}(q(x))
\end{equation}
Typically, a flow-based model transforms the covariates to a latent space with a simple base distribution, like a standard Gaussian. However, TANs provide additional flexibility by modeling the latent distribution with an autoregressive approach \cite{larochelle2011neural}. This alters the earlier equation \eqref{eq:changevariables}, in that $p_{\mathcal{Z}}(q(x))$ is now represented as the product of $d$ conditional distributions.
\begin{equation}\label{eq:autoreg}
    p_{\mathcal{X}}(x) = \left| \det \frac{dq}{dx} \right| \prod_{i=1}^d p_{\mathcal{Z}}(z^i \mid z^{i-1},...,z^{1})
\end{equation}
Since flow models give the exact likelihood, they can be trained by directly optimizing the log likelihood. In addition, thanks to the invertibility of the transformations, one can draw samples by simply inverting the transformations over a set of samples from the latent space.

\section{Methods}

\begin{figure*}
    \centering
    \subfigure[general formulation]{
    \label{fig:gen_trans}
    \includegraphics[width=0.25\linewidth]{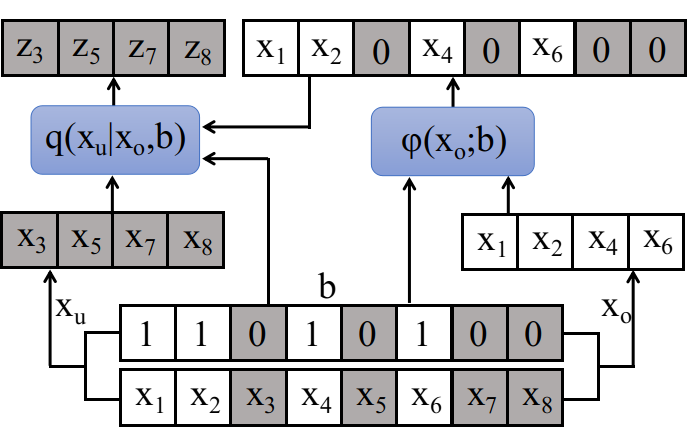}
    }
    \subfigure[affine coupling]{
    \label{fig:affine}
    \includegraphics[width=0.21\linewidth]{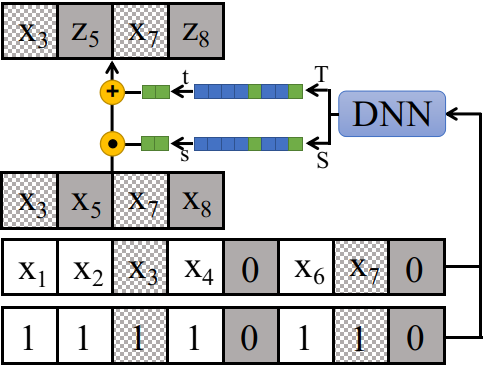}
    }
    \subfigure[linear]{
    \label{fig:linear}
    \includegraphics[width=0.23\linewidth]{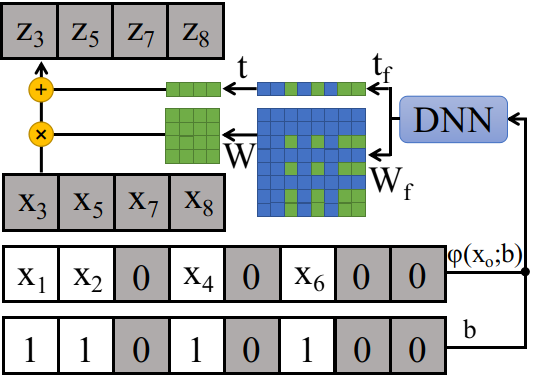}
    }
    \subfigure[RNN coupling]{
    \label{fig:rnncoupling}
    \includegraphics[width=0.24\linewidth]{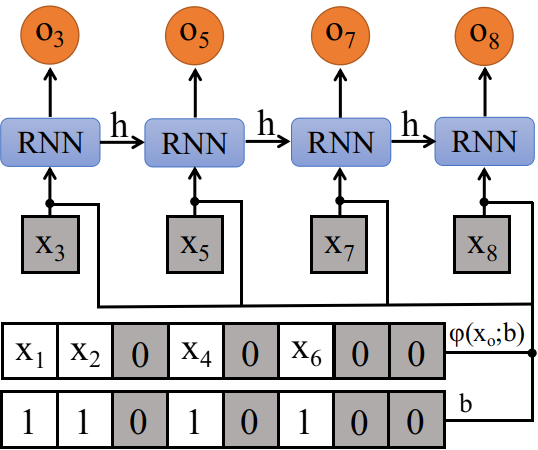}
    }
    \caption{Conditional transformations used in \method. Grayed out boxes represent unobserved covariates. Checkerboarded boxes in (b) belong to unobserved dimensions, but are used as conditioning in the affine coupling transformation.}
\end{figure*}

We develop \method by constructing both conditional transformations of variables and autoregressive likelihoods that work with an arbitrary set of unobserved covariates.
To deal with arbitrary dimensionality for conditioning covariates $x_o$, we define a zero imputing function $\phi(x_o;b)$ that returns a $d$-dimensional vector by imputing vector $x_o \in \mathbb{R}^{|o|}$ with zeros based on the specified binary mask $b$:
\begin{equation}\label{eq:zeroimputation}
    w = \phi(x_o;b),\quad
    w_i = \left\{
            \begin{array}{rcl}
            x_o[c_i], & &b_i = 1\\
            0,   & &b_i = 0
            \end{array}
          \right.
\end{equation}
where $c_i=\sum_{j=1}^{i}b_j$ represents the cumulative sum over $b$. The imputed output $w$ is a $d$-dimensional vector with unobserved values replaced by zeros (See Fig.~\ref{fig:gen_trans} for an illustration.). Thus, we get a conditioning vector with fixed dimensionality.
However, handling the arbitrary dimensionality of $x_u$ requires further care, as discussed below.

\subsection{Arbitrary Conditional Transformations}
We first consider a conditional extension to the change of variable theorem:
\begin{equation}\label{eq:condtrans}
    p_{\mathcal{X}}(x_u \mid x_o, b) = \left| \det \frac{dq_{x_o,b}}{dx_u} \right| p_{\mathcal{Z}}(q_{x_o,b}(x_u) \mid x_o,b),
\end{equation}
where $q_{x_o, b}$ is a transformation on the unobserved covariates $x_u$ with respect to the observed covariates $x_o$ and binary mask $b$ as demonstrated in Fig.~\ref{fig:gen_trans}.
However, the fact that $x_u$ is a set of arbitrary missing dimensions makes it challenging to define $q_{x_o, b}$'s across different bitmasks $b$.
One challenge comes from requiring the transformation to have adaptive outputs that can adapt to different dimensionality of $x_u$. Another challenge is that different missing patterns require the transformation to capture different dependencies. Since the missing pattern could be arbitrary, we require the transformation to learn a large range of possible dependencies.
To solve those challenges, we propose several conditional transformations that leverage the conditioning information in $x_o$ and $b$ and can be adapted to arbitrary $x_u$. We describe them in detail below.
For notation simplicity, we drop the subscripts $\mathcal{X}$ and $\mathcal{Z}$ in the following sections.

\paragraph{Affine Coupling Transformation}
Affine coupling is a commonly used flow transformation. Here, we derive a arbitrary conditional version.
Just as in the unconditional counterparts \cite{dinh2014nice,dinh2016density,kingma2018glow}, we divide the unobserved covariates $x_u$ into two parts, $x_u^A$ and $x_u^B$ according to some binary mask $b_u \in \{0,1\}^{|u|}$, i.e., $x_u^A=x_u[b_u]$ and $x_u^B=x_u[1-b_u]$. We then keep the first part $x_u^A$ and transform the second part $x_u^B$, i.e.,
\begin{align}
\begin{split}
z_u^A &= x_u^A\\
z_u^B &= x_u^B \odot s(x_u^A, b_u, x_o, b)+t(x_u^A, b_u, x_o, b),
\end{split}
\end{align}
where $\odot$ represents the element-wise (Hadamard) product. $s$ and $t$ computes the scale and shift factors as the function of both observed covariates and covariates in group A.
Note that both inputs and outputs of $s$ and $t$ contain arbitrary dimensional vectors. To deal with arbitrary dimensionality in inputs, we apply the zero imputing function \eqref{eq:zeroimputation} to $x_u^A$ and $x_o$ respectively to get two $d$-dimensional vectors with missing values imputed by zeros. We also apply $\phi$ to $b_u$ to get a $d$-dimensional mask. 
The shift and scale functions $s$ and $t$ are implemented as deep neural networks (DNN) over $x_{c} =\phi(\phi(x_u^A;b_u);1-b) + \phi(x_o;b)$ and $b_{c} = \phi(b_u;1-b) + b$, i.e,
\begin{align}
\begin{split}
    S &= \text{DNN} (\mathbf{concat}(x_c, b_c))\in\mathbb{R}^d,\\
    T &= \text{DNN} (\mathbf{concat}(x_c, b_c))\in\mathbb{R}^d,
\end{split}    
\end{align}
where $\mathbf{concat}$ defines a concatenate function and the two DNN functions can share weights.

The outputs of $s$ and $t$ need to be adaptive to the dimensions of $x_u^B$, thus we apply the indexing mechanism, $[\cdot]$, that takes the corresponding dimensions of non-zero values with respect to binary masks $1-b$ and $1-b_u$, i.e.,
\begin{align}
\begin{split}
    s &= S\left[1-b\right]\left[1-b_u\right],\\
    t &= T\left[1-b\right]\left[1-b_u\right].
\end{split}    
\end{align}
A visualization of this transformation is presented in Fig.~\ref{fig:affine}.
Note that the way we divide $x_u$ might depend on our prior knowledge about the data correlations. For image data, we use checkerboard and channel-wise split as in \cite{dinh2016density}. For real-valued vectors, we use even-odd split as in \cite{dinh2014nice}.

\paragraph{Linear Transformation}
Another common transformation for flow-based models is the linear transformation. Contrary to the coupling transformation that only leverages correlation between two separated subsets, the linear transformation can take advantage of correlation between all dimensions. Furthermore, the linear transformation can be viewed as a generalized permutation which rearranges dimensions so that next transformation can be more effective. 

In order to transform $x_u$ linearly, we would like to have an adaptive weight matrix $\mathbf{W}$ of size $\left|u\right| \times \left|u\right|$ and a bias vector $t$ of size $|u|$. Similar to the affine coupling described above, we first apply a deep neural network over $\phi(x_o;b)$ and binary mask $b$ to get a $d \times d$ matrix $\mathbf{W_f}$ and a $d$-dimensional bias vector $t_f$, then we index them with respect to the binary mask $1-b$, i.e.,
\begin{align}
\begin{split}
    \mathbf{W_f} &= \text{DNN}(\phi(x_o;b), b) \in\mathbb{R}^{d\times d},\\
    \mathbf{W} &= \mathbf{W_f}\left[1-b, 1-b\right]\in\mathbb{R}^{|u|\times |u|},\\
    t_f &= \text{DNN}(\phi(x_o;b), b) \in\mathbb{R}^{d},\\
    t &= t_f\left[1-b\right]\in\mathbb{R}^{|u|}.
\end{split}
\end{align}
The linear transformation can then be derived as $z_u = \mathbf{W}x_u + t$. Fig.~\ref{fig:linear} illustrates this transformation over an 8-dimensional example. In practice, we add another learnable full-rank weight matrix to $W_f$ to guarantee invertibility.

In order to decrease complexity, it is straightforward to parametrize $W_f$ with rank $r$ matrices by taking the product of two rank $r$ matrices with size $d \times r$ and $r \times d$ respectively. Hence, the DNN can reduce its output dimensions to $2dr$. During preliminary experiments, we observed minimal drop in performance when using a large enough $r$.

\paragraph{RNN Coupling Transformation}
The affine coupling transformation can be viewed as a rather rough recurrent transformation with only one recurrent step.
We can generalize it by running an RNN over $x_u$ and transform each dimension sequentially (shown in Fig.\ref{fig:rnncoupling}). Note that a recurrent transformation naturally handles different dimensionality. 
To leverage conditioning inputs $x_o$ and $b$, we concatenate $\phi(x_o;b)$ and $b$ to each dimension of $x_u$.
The outputs of the RNN are used to derive the shift and scale parameters respectively.
\begin{align}
\begin{split}
    o^i, h^i &= \text{RNN}(\mathbf{concat}(x_u^{i-1},\phi(x_o;b),b), h^{i-1}),\\
    z_u^i &= x_u^i * s(o^i) + t(o^i),\\
\end{split}
\end{align}
where $x_u^0 = \mathbf{-1}$, $h^0 = \mathbf{0}$,
and $i$ indexes through dimensions.

\paragraph{Other Transformations and Compositions} Other transformations like element-wise leaky-ReLU transformation \cite{oliva2018transformation} and reverse transformation \cite{dinh2014nice} are  readily applicable to transform the unobserved covariates $x_u$ since they do not rely on the conditioning covariates.
Other than these specific transformations described above, any transformations that follow the general formulation shown in Fig.~\ref{fig:gen_trans} can be easily plugged into our model. 
We obtain flexible, highly non-linear transformations with the composition of multiple of these aforementioned transformations.
(That is, we use the output of the preceding transformation as input to the next transformation.)
The Jacobian of the resulting composed (stacked) transformation is accounted with the chain rule.

\subsection{Arbitrary Conditional Likelihoods}
The conditional likelihoods in latent space $p(z_u | x_o,b)$ can be computed by either a base distribution, like a Gaussian, or an autoregressive model as in TANs. For Gaussian based likelihoods, we can get mean and covariance parameters by applying another function over $\phi(x_o;b)$ and $b$, which is essentially equivalent to another linear transformation conditioned on $x_o$ and $b$. However, this approach is generally less flexible than using an autoregressive approach.

For autoregressive likelihoods, conditioning vectors can be used in the same way as the RNN coupling transformation. The difference is that the RNN outputs are now used to derive the parameters for some base distribution, for example, a Gaussian Mixture Model:
\begin{align}
\begin{split}
    o^i, h^i = \text{RNN}(\mathbf{concat}(z_u^{i-1},\phi(x_o;b),b), h^{i-1}),\\
    p(z_u^i \mid z_u^{i-1}, ..., z_u^1, x_o, b) = \text{GMM}(z_u^i \mid \theta(o^i)),\\
\end{split}
\end{align}
where $\theta(o^i)$ is a shared fully connected network that maps to the parameters for the mixture model (i.e. each mixture component's location, scale, and mixing weight parameter). During sampling, we iteratively sample each point, $z_u^{i-1}$, before computing the parameters for $z_u^i$. Incorporating the autoregressive likelihood into Eq. \eqref{eq:condtrans} yields:
\begin{equation}\label{eq:condtans}
    p(x_u \mid x_o, b) = \left| \det \frac{dq_{x_o,b}}{dx_u} \right| \prod_{i=1}^{|u|} p(z_u^i \mid z_u^{i-1},...,z_u^{1}, x_o, b),
\end{equation}
where $|u|$ is the cardinality of the unobserved covariates.

\subsection{Missing Data}

During training, if we have access to complete training data, we will need to manually create binary masks $b$ based on some predefined distribution $p_b$. $p_b$ is typically chosen based on the application. For instance, Bernoulli random masks are commonly used for real-valued vectors. Given binary masks, training data $x$ are divided into $x_u$ and $x_o$ and fed into the conditional model $p(x_u \mid x_o,b)$.

If training data already contains missing values, we can only train our model on the remaining covariates. As before, we manually split each data point into two parts, $x_u$ and $x_o$ based on a binary mask $b$. Note that dimensions in $b$ corresponding to the missing values are always set to $0$, i.e., they are never observed during training. In this setting, we will need another binary mask $m$ indicating those dimensions that are not missing. Accordingly, we define observed dimensions as $x_o=x\left[b\right]$ and unobserved dimensions as $x_u=x\left[m \odot (1-b)\right]$ and optimize $p(x_u \mid x_o,m,b)$. 

\subsection{Special Cases}
We can easily see that arbitrary conditional model $p(x_u \mid x_o, b)$ is a special case of $p(x_u \mid x_o, m, b)$ if we set all dimensions of the binary mask $m$ to one (no missing data). 
As a special case of \method, the joint distribution $p(x)$ can be modeled by the same framework when we set $m$ to all ones (no missing data) and $b$ to all zeros (no observed dimensions). Essentially, we are trying to model all the dimensions by conditioning on an empty set $p(x \mid \emptyset)$. Similarly, we can use \method to model the arbitrary marginal distribution $p(x_u)$ by setting $m = 1-b$, that is, all the observed dimensions are treated as missing data. Note that $x_u$ is an arbitrary subset of the covariates, thus, we are essentially modeling a mixture of all the marginals in just one single model.

\subsection{Imputation and Best Guess Objective}
Given a trained \method model, multiple imputations can be easily accomplished by drawing multiple samples from the learned conditional distribution.
However, certain downstream tasks may require a single imputed ``best guess''. Unfortunately, the analytical mean $\mathbb{E}_{p(x_u \mid x_o,b)}[x_u]$  is not available for flow-based deep generative models. Furthermore, getting an accurate empirical estimate could be prohibitive in high dimensional space.
In this work, we propose a robust solution that gives a single best guess in terms of the MSE metric (it can be easily extended to other metrics, e.g. an adversarial one).

Specifically, we obtain our best guess by inverting the conditional transformation over the mean of the latent distribution, i.e.,
\begin{equation}
q_{x_o,b}^{-1}(\bar{z})=q_{x_o,b}^{-1}(\mathbb{E}_{p_{\mathcal{Z}}(z \mid x_o,b)}[z]).
\end{equation}
The mean $\bar{z}$ is analytically available for Gaussian mixture base model.
To ensure that this best guess is close to unobserved values, we optimize with an auxiliary MSE loss:
\begin{equation}
    \mathcal{L} = -\log~p(x_u \mid x_o, b) + \lambda \|q_{x_o,b}^{-1}(\bar{z}) - x_u \|^2,
\end{equation}
where $\lambda$ controls the relative importance of the auxiliary objective.
Note that we only penalize one particular point from $p(x_u \mid x_o,b)$ to be close to $x_u$. Hence, it does not affect the diversity of the conditional distribution.

\section{Related Work}
\subsection{Arbitrary Conditional Models}
Previous attempts to learn probability distributions conditioned on arbitrary subsets of known covariates include the Universal Marginalizer  \cite{douglas2017marginalizer}, which is trained as a feed-forward network to approximate the marginal posterior distribution of each unobserved dimension conditioned on the observed ones. 
VAEAC \cite{ivanov2018variational}, the \emph{state-of-the-art} model so far for modeling arbitrary conditional likelihoods, utilizes a conditional variational autoencoder and extends it to deal with arbitrary conditioning. The decoder network outputs likelihoods that are over all possible dimensions, although, since they are conditionally independent given the latent code, it is possible to use only the likelihoods corresponding to the unobserved dimensions.
NeuralConditioner (NC) \cite{belghazi2019learning} is a GAN-based approach which leverages a discriminator to distinguish real data and generated samples.

Unlike VAEAC and NC, \method is capable of producing an analytical (normalized) likelihood and avoids blurry samples and mode collapse problems ingrained in these approaches. Furthermore, in contrast to the Universal Marginalizer, \method captures dependencies between unobserved covariates at training time via the change of variables formula.

Another type of model, Sum-Product Network (SPN) \cite{poon2011sum}, where the network structures are specially designed and contain only sum and product operations can evaluate both arbitrary conditional likelihoods and marginal likelihoods efficiently. SPN builds a DAG by stacking sum and product operations alternately so that the partition function can be efficiently computed. In contrast, ACFlow is more flexible since we do not pose constraints on the network structures. 
Please see Sec.~\ref{sec:spn} for empirical comparison between ACFlow and SPNs.

\subsection{Missing Data Imputation}
Classic methods for imputation include $k$-nearest neighbors \cite{troyanskaya2001missing}, random forest \cite{stekhoven2011missforest} and auto-encoder \cite{gondara2018mida} approaches. 
Deep generative models have already been explored to handle missing data. MIWAE \cite{mattei2019miwae} train a VAE model by a modified lower bound tailored for missing data and perform imputation by importance sampling.
GAIN \cite{yoon2018gain} addresses data imputation with a GAN approach, where a generator produces imputations and a discriminator attempts to distinguish imputed covariates from observed covariates. MisGAN \cite{li2019misgan} is yet another GAN based method where they train a generator-discriminator pair for both data and masks.

Many missing data imputation methods assume that the data are missing completely at random (MCAR, missing independently of covariates' values) or missing at random (MAR, possibly missing depending on observed covariates' values) \cite{little2019statistical}. Missing not at random (MNAR, missing depending on unobserved covariates' values) is much harder to address, and requires some approximate inference technique such as variational approaches \cite{murray2018multiple}. In this work, we focus on the MCAR scenario.

\section{Experiments}\label{experiments}

\subsection{Synthetic Datasets}

\begin{figure}[htb]
    \centering
    \includegraphics[width=0.95\linewidth]{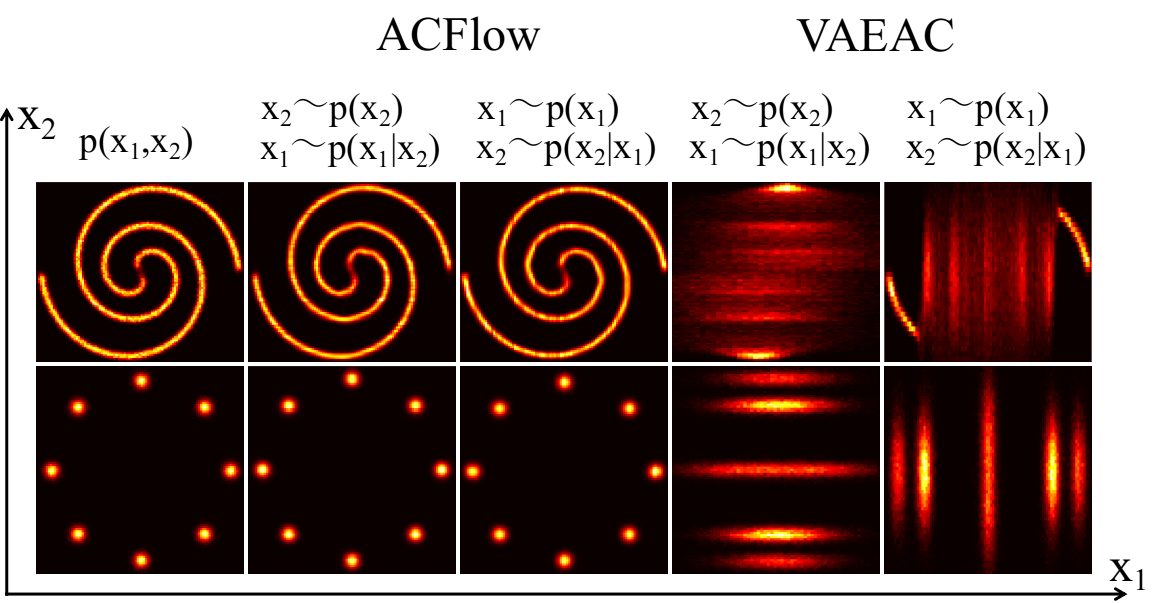}
    \caption{Synthetic datasets. Observed covariates are sampled from the marginal distributions, and the missing covariates are sampled from the learned conditional distributions.}
    \label{fig:syn}
\end{figure}

To validate the effectiveness of our model, we conduct experiments on synthetic 2-dimensional datasets, i.e., $x=(x_1,x_2) \in \mathbb{R}^2$. The joint distributions and imputed samples are plotted in Fig.~\ref{fig:syn}. 
Here, the conditional distributions are highly multi-modal and vary significantly when conditioned on different observations. We train arbitrary conditional models on 100,000 samples from the joint distribution. The masks are generated by dropping out one dimension at random.

We compare our model against VAEAC \cite{ivanov2018variational}, the current \emph{state-of-the-art} model trained purely by likelihood. 
We closely follow the released code for VAEAC\footnote{\url{https://github.com/tigvarts/vaeac/blob/master/imputation_networks.py}}
to construct the proposal, prior, and generative networks. Specifically, we use fully connected layers and skip connections as is in their official implementation. We also use short-cut connections between the prior network and the generative network. We search over different combinations of the number of layers, the number of hidden units of fully connected layers, and the dimension of the latent code. Validation likelihoods are used to select the best model.
The details about training procedure are provided in Appendix \ref{sec:syn}.
We see that \method is capable of learning multi-modal distributions, while VAEAC tends to merge multiple modes into one single mode.

\begin{figure*}[htb]
    \centering
    \includegraphics[width=0.95\linewidth]{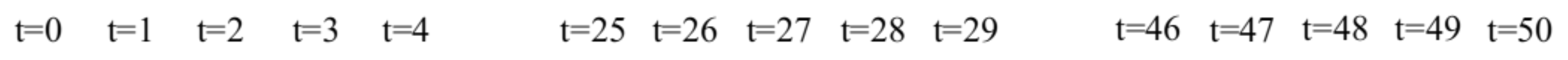}
    \includegraphics[width=0.95\linewidth]{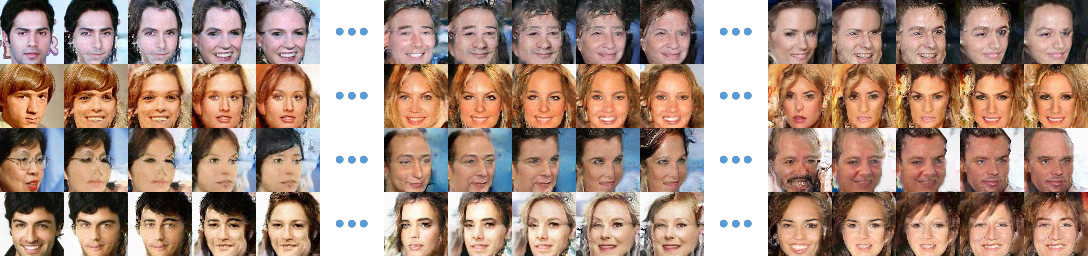}
    \caption{Gibbs sampling using a trained \method. We iteratively sample the upper or lower half conditioned on the rest for 50 mixing steps. Note we reduce the bit depth to 5 in this experiment for better sample quality. More samples are shown in Appendix~\ref{sec:gibbs}.}
    \label{fig:gibbs}
\end{figure*}

\subsection{Likelihood Evaluation}
In this section, we evaluate the ability of \method to model arbitrary conditional distributions in terms of the negative log-likelihoods (NLL). We compare to VAEAC \cite{ivanov2018variational} and an autoregressive counterpart \cite{salimans2017pixelcnn++,larochelle2011neural}. In addition, as special cases, we compare arbitrary marginal likelihoods and the joint likelihoods against flow models. 

\subsubsection{Image Datasets}

We evaluate our method on three common image datasets: MNIST, Omniglot and CelebA. We assume access to complete training data and train arbitrary conditional models $p(x_u \mid x_o, b)$ with a varied distribution of binary masks $p_b$. Details about mask generation and image preprocessing are available in the Appendix \ref{sec:inpaint_data}.

The architecture of \method contains a sequence of arbitrary conditional transformations, which is akin to the RealNVP model \cite{dinh2016density} except we replace all the coupling layers with our proposed arbitrary conditional alternative. The arbitrary conditional likelihood is implemented as a conditional PixelCNN conditioned on a vector embedding of the observed covariates. Implementation details of \method and baselines are provided in Appendix \ref{sec:inpaint_model}.

\begin{table}[htb]
    \caption{NLL for modeling $p(x_u \mid x_o)$. Mean and std. dev. are computed by sampling 5 binary masks at random for each testing data. Lower is better.}
    \label{tab:inpaint_nll}
    \centering
    \scriptsize
    \begin{tabular}{cccc}
    \toprule
         & MNIST & Omniglot & CelebA  \\
    \midrule
    VAEAC & 1034.86\tabpm6.05 & 1014.62\tabpm2.53 & 17638.83\tabpm15.21\\
    PixelCNN & 313.93\tabpm3.09 & 160.20\tabpm1.97 & 8806.02\tabpm34.95\\
    ACFlow & \textbf{271.99\tabpm3.10} & \textbf{159.32\tabpm1.25} & \textbf{8387.01\tabpm45.86}\\
    \bottomrule
    \end{tabular}
\end{table}

We first evaluate the ability to model arbitrary conditional distributions by comparing the negative log-likelihoods (NLL) in Table.~\ref{tab:inpaint_nll}. We generate 5 different masks for each test image and report the average scores and the standard deviation. We see that \method outperforms VAEAC by a large margin and achieve \emph{state-of-the-art} performance in terms of arbitrary conditional likelihoods. \method also outperforms the PixelCNN model due to the transformations' ability to capture dependencies across all covariates, while a PixelCNN model can only leverage preceding covariates to inference the current covariate.

\begin{figure*}[]
    \centering
    \includegraphics[height=3.1cm]{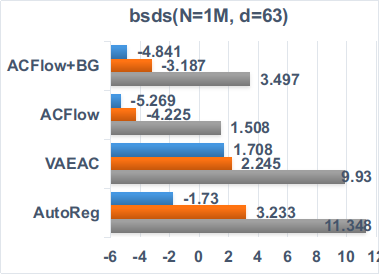}
    \includegraphics[height=3.1cm]{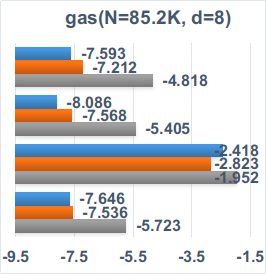}
    \includegraphics[height=3.1cm]{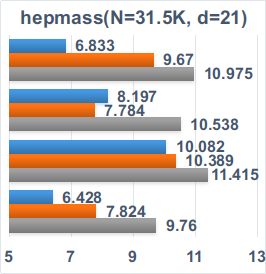}
    \includegraphics[height=3.1cm]{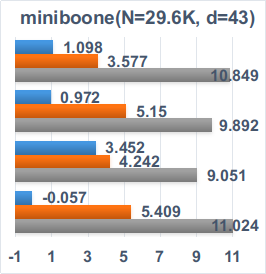}
    \includegraphics[height=3.1cm]{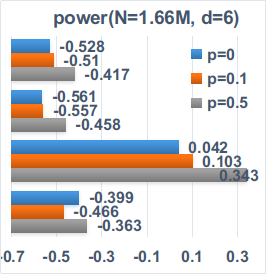}
    \caption{NLL for models trained on different level of missing rates (p). Lower is better. On top of each chart, we present the dataset name and its training set size (N) and feature dimensions (d). ``BG" indicates the proposed best guess penalty.}
    \label{fig:uci_nll}
\end{figure*}

As a special case, we evaluate the joint likelihood $p(x)$ by setting the binary mask to all zeros during testing, i.e. $p(x \mid \emptyset)$, where we condition on an empty set to get likelihood for all covariates. Note that the model is trained using the conditional likelihood $p(x_u \mid x_o)$ rather than the joint likelihood.
\begin{wraptable}{r}{4cm}
    \caption{Evaluate joint likelihood $p(x)$ by setting bitmask $b$ to all zeros. Results are presented as bits per dimension (bpd). Lower is better.}
    \label{tab:inpaint_joint}
    \centering
    \scriptsize
    \begin{tabular}{ccc}
    \toprule
           &  ACFlow & RealNVP \\
    \midrule
    MNIST & 0.86 & 0.87\\
    Omniglot & 0.52 & 0.52\\
    CIFAR10 & 3.50 & 3.49 \\
    CelebA & 2.85 & 3.02 \\
    \bottomrule
    \end{tabular}
\end{wraptable}
\normalsize
In Table.~\ref{tab:inpaint_joint} we compare to the RealNVP model with similar architecture. We use Gaussian likelihood in this experiment for fair comparison. We see ACFlow achieves better or comparable likelihoods despite not directly trained to model the joint distribution. We believe that training ACFlow for multiple related tasks (the conditional likelihood tasks) with tied weights may act as a regularizer.
Samples from the joint distributions are shown in Appendix~\ref{sec:joint}.

Besides conditioning on an empty set, we can apply a Gibbs sampling procedure to sample from the joint distribution. As shown in Fig.~\ref{fig:gibbs}, we iteratively sample the upper or lower half conditioning on the remaining half for 50 mixing steps. We can see that the image changes smoothly and mixes well as the samples we get are very different from the starting point. This Gibbs sampling can be viewed as a way of exploring the local manifold from a specified starting point, an application that we will explore further in future work.

\subsubsection{Real-valued Datasets}

Next, we evaluate our model on real-valued tabular datasets.
We use UCI repository datasets preprocessed as described in \cite{papamakarios2017masked}.
We construct models by composing several leaky-ReLU, conditional linear, and RNN coupling transformations, along with an autoregressive arbitrary conditional likelihood component. Please refer to Appendix.~\ref{sec:imputation_model} for further architectural details.

First, we consider non-missing training data. We construct masks, $b$, by dropping a random subset of the dimensions according to a Bernoulli distribution with $p=0.5$.
Afterwards, we also evaluate our model when the training data itself contains missing values that are never available during training. We consider training and testing with data features missing completely at random at a 10\% and 50\% rate.

Figure.~\ref{fig:uci_nll} presents the arbitrary conditional NLL for models trained with different level of missing rates. We compare to the current state-of-the-art, VAEAC, and an autoregressive model. One can see that our model gives better NLL on nearly all scenarios compared to VAEAC, which indicates our model is better at learning the true distribution. Our model also outperforms the autoregressive model for most cases.

\begin{table}[]
    \caption{Marginal log-likelihood evaluated on the first $d$ dimensions. Note that \method captures all the marginals in one single model, while different TANs are trained specifically for each case. Higher is better.}
    \label{tab:uci_margin}
    \centering
    \scriptsize
    \begin{tabular}{ccc|cc|cc}
    \toprule
         & \multicolumn{2}{c|}{d=3} & \multicolumn{2}{c|}{d=5} & \multicolumn{2}{c}{d=10} \\
         & \method & TAN & \method & TAN & \method & TAN \\
    \midrule
    bsds & 5.06 & 5.11 & 9.26 & 9.43 & 19.60 & 20.44 \\
    gas & 0.78 & 1.22 & 3.01 & 4.47 & 10.13 & 12.09  \\
    hepmass & -4.03 & -4.00 & -6.19 & -5.92 & -11.58 & -10.87 \\
    miniboone & -2.76 & -2.13 & -5.31 & -3.73 & -10.36 & -8.13 \\
    power & -0.57 & -0.54 & 1.34 & 1.40 & 0.42 & 0.57 \\
    \bottomrule
    \end{tabular}
\end{table}
\normalsize

In addition to the arbitrary conditional likelihoods, we can also train \method to learn arbitrary marginal likelihoods. Similar to the arbitrary conditional case, a set of covariates have an exponential number of marginal distributions. We tested our model against TAN models trained explicitly on a particular subset of covariates (we use the first $d$ dimensions for convenience); results are shown in Table.~\ref{tab:uci_margin}.
Scatter plots from the first 3 dimensions of miniboone are shown in Fig.~\ref{fig:marg} and samples of other datasets are available in Appendix Fig.~\ref{fig:margin}.
\begin{wrapfigure}{r}{4cm}
    \centering
    \vspace{-5pt}
    \includegraphics[width=\linewidth,trim=2.4cm 0.4cm 2.4cm 1.6cm,clip]{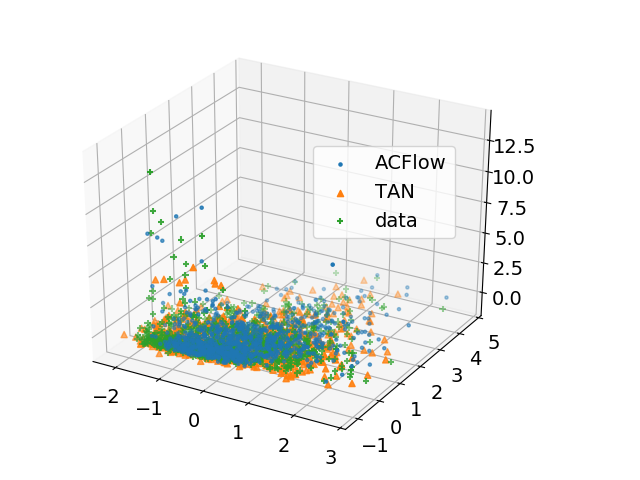}
    \caption{Marginal samples of miniboone from the first 3 dimensions.}
    \label{fig:marg}
\end{wrapfigure}
Although the ACFlow model was trained for arbitrary marginals (and did not know which marginal tasks it would be evaluated on), it performed as well as flow models that were traned specifically for these marginals. Thus, we expect that a single ACFlow model would also generalize to any arbitrary marginal task, which would require one to train an exponential number of TAN flow models (one for each task). In essence, ACFlow is able to do approximate marginalization for flow methods, which was previously intractable.

\subsubsection{Comparison with SPNs}\label{sec:spn}

SPNs \cite{poon2011sum} model joint distributions with a specially designed architecture so that both arbitrary conditionals and marginals are tractable.
\begin{wraptable}{l}{4cm}
    \centering
    \vspace{-5pt}
    \caption{Comparison with DCSPN on Olivetti Face dataset. MSE scores for DCSPN are from \cite{butz2019deep}. Lower is better.}
    \label{tab:dcspn}
    \footnotesize
    \begin{tabular}{ccc}
        \toprule
         &  left & bottom \\
        \midrule
        DCSPN & 455 & 503 \\
        ACFlow & 415 & 434 \\
        \bottomrule
    \end{tabular}
\end{wraptable}
First, we compare to the DCSPN \cite{butz2019deep} on Olivetti Face dataset. We evaluate the imputation performance by sampling from the conditional distributions $p(x_u \mid x_o)$ using two different masks (denoted as `left' and `bottom'). Results are shown in Tab.~\ref{tab:dcspn}. The MSE scores for DCSPN are from their paper. Note that they train two separate models for different masks, while we handle both cases in one single ACFlow model and still obtain lower MSE scores.

Next, we conduct experiments on UCI dataset to compare with SPFlow \cite{molina2019spflow}, a python library for SPNs. We report results in terms of the conditional NLL, the NRMSE, and the marginal NLL in Tab.~\ref{tab:spflow}. The marginal likelihoods are evaluated for the first 3, 5 and 10 dimensions, respectively. In most of the cases, ACFlow outperforms SPFlow.

\begin{table}[h!]
    \centering
    \caption{Comparison to SPFlow on UCI datasets. Lower is better.}
    \label{tab:spflow}
    \scriptsize
    \resizebox{\linewidth}{!}{
    \begin{tabular}{ccccccc}
    \toprule
         &  & bsds & gas & hepmass & miniboone & power \\
    \midrule
       \multirow{2}{*}{$p(x_u \mid x_o)$}  & SPFlow & 24.15 & -4.30 & 12.78 & 18.34 & 1.03\\
                                           & ACFlow &-5.27 & -8.09 &  8.20 & 0.97 & -0.56\\
    \midrule
       \multirow{2}{*}{NRMSE} & SPFlow & 1.01 & 0.37 & 1.02 & 0.76 & 0.95\\
                              & ACFlow & 0.60 & 0.57 
                              & 0.91 & 0.48 & 0.88\\
    \midrule
       \multirow{2}{*}{$p(x[:3])$} & SPFlow & 2.87 & -0.68 &  4.01         & 2.21 &  0.63 \\
                                   & ACFlow & -5.06 & -0.78 &  4.03     &  2.76 & 0.57\\
    \midrule
       \multirow{2}{*}{$p(x[:5])$} & SPFlow & 4.42 & -1.88 &  6.58         &   4.31 &  -1.01\\
                                   & ACFlow &  -9.26 & -3.01 &  6.19         &  5.31 & -1.34\\
    \midrule
       \multirow{2}{*}{$p(x[:10])$} & SPFlow & 8.15 & -4.81 & 13.38 & 9.85 & 0.12\\
                                   & ACFlow & -19.60 & -10.13 & 11.58 & 10.36 & -0.42\\
    \bottomrule
    \end{tabular}}
\end{table}

\subsection{Imputation}

Applying \method for data imputation is straight-forward, since we can sample from the learned conditional distributions $p(x_u \mid x_o)$. In this section, we evaluate the performance of multiple and single imputations against other likelihood based generative models, such as VAEAC and autoregressive models. 
We also compare to classic imputation methods, such as MICE \cite{micebuuren2010} and MissForest \cite{missstekhoven2012}.

\subsubsection{Image Inpainting}

\begin{figure}[]
    \centering
    \subfigure[MNIST inpaintings]{
    \includegraphics[width=0.46\linewidth]{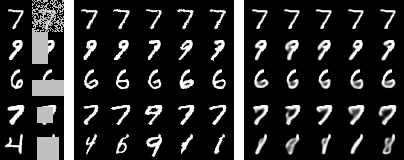}
    }
    \subfigure[Omniglot inpaintings]{
    \includegraphics[width=0.46\linewidth]{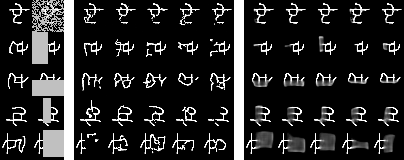}
    }
    \subfigure[CelebA inpaintings]{
    \includegraphics[width=0.96\linewidth]{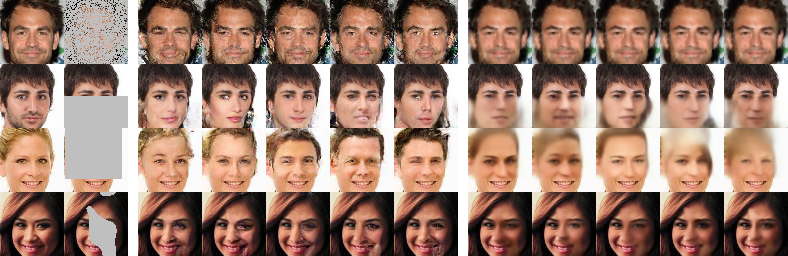}
    }
    \caption{Image Inpaintings. Left: groundtruth and inputs. Middle: samples from \method. Right: samples from VAEAC.}
    \label{fig:inpaint}
\end{figure}

\begin{figure*}[]
    \centering
    \includegraphics[height=4.8cm]{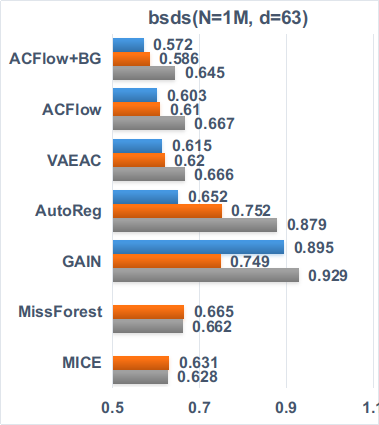}
    \includegraphics[height=4.8cm]{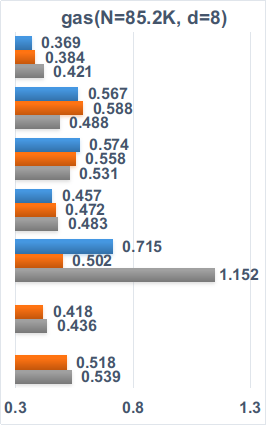}
    \includegraphics[height=4.8cm]{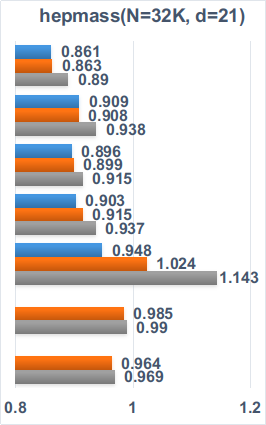}
    \includegraphics[height=4.8cm]{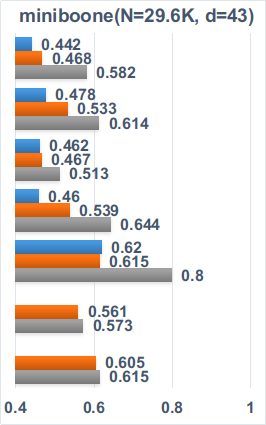}
    \includegraphics[height=4.8cm]{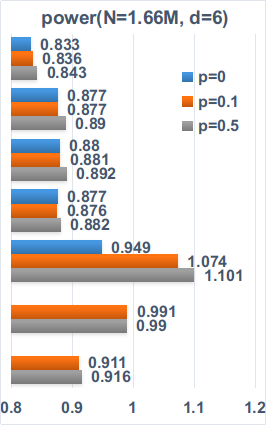}
    \caption{NRMSE results for real-valued feature imputation on UCI datasets. Lower is better. We test MICE and MissForest only when the missing rate (p) is not zero.}
    \label{fig:uci_mse}
\end{figure*}

Figure.~\ref{fig:inpaint} shows multiple imputation results from VAEAC and \method (We use a Gaussian base likelihood for this experiment, since sampling from an autoregressive model is time consuming.). More samples are available in Appendix \ref{sec:inpaint_sample}. We notice \method can generate coherent and diverse inpaintings for all three datasets and different masks. Compared to VAEAC, our model generates sharp samples and restores more detail. Even when the missing rate is high, \method can still generate decent inpaintings.

To quantitatively evaluate the inpainting performance, we report the peak signal-to-noise ratio (PSNR) and the precision and recall scores (PRD) \cite{sajjadi2018assessing} in Table.~\ref{tab:inpaint}. We note that PSNR is a metric that may prefer blurry images over sample diversity \cite{5596999}, hence we evaluate the trade-off between sample quality and diversity via the precision and recall scores (PRD) \cite{sajjadi2018assessing}. Since we cannot sample from the groundtruth conditional distribution, we compute the PRD score between the imputed joint distribution $p(x_o)p(x_u \mid x_o)$ and the true joint distribution $p(x)$ via sampling 10,000 points. The PRD scores for two distributions measure how much of one distribution can be generated by another. Higher recall means a greater portion of samples from the true distribution $p(x)$ are covered by $p(x_o)p(x_u \mid x_o)$; and similarly, higher precision means a greater portion of samples from $p(x_o)p(x_u \mid x_o)$ are covered by $p(x)$. We report the $(F_8, F_{\frac{1}{8}})$ pairs in Table.~\ref{tab:inpaint} to represent recall and precision, respectively.

From the quantitative results, we see that our model gives higher PSNR and PRD scores compared to VAEAC, demonstrating that our model better learns the true distribution.
Training with the auxiliary ``best guess'' penalty (denoted as ``ACFlow+BG'') can further improve the PSNR scores significantly, but hardly impacts the PRD scores, which verifies that the proposed penalty does not affect the diversity of the model.


\begin{table}[]
    \caption{Inpainting results. Mean and std. dev. are computed by sampling 5 binary masks for each testing data. Higher is better.}
    \label{tab:inpaint}
    \centering
    \scriptsize
    \resizebox{\linewidth}{!}{
    \begin{tabular}{ccccc}
    \toprule
         & & MNIST & Omniglot & CelebA \\
    \midrule
    \multirow{2}{*}{VAEAC}
    & PSNR & 19.613\tabpm0.042 & 17.693\tabpm0.023 & 23.656\tabpm0.027 \\ \cmidrule{2-2}
    & PRD & (0.975, 0.877) & (0.926, 0.525) & (0.966, 0.967) \\
    \midrule
    \multirow{2}{*}{ACFlow}   
    & PSNR & 17.349\tabpm0.018 & 15.572\tabpm0.031 & 22.393\tabpm0.040 \\ \cmidrule{2-2}
    & PRD  & (\textbf{0.984}, 0.945) & (\textbf{0.971}, 0.962) & (\textbf{0.988}, \textbf{0.970}) \\
    \midrule
    \multirow{2}{*}{ACFlow+BG}   
    & PSNR & \textbf{20.828}\tabpm\textbf{0.031} & \textbf{18.838}\tabpm\textbf{0.009} & \textbf{25.723}\tabpm\textbf{0.020} \\ \cmidrule{2-2}
    & PRD & (0.983, \textbf{0.947}) & (0.970, \textbf{0.967}) & (0.987,0.964)\\
    \bottomrule
    \end{tabular}}
\end{table}

\subsubsection{Feature Imputation}

In Figure.~\ref{fig:uci_mse}, we compare feature imputation performance using NRMSE (i.e. root mean squared error normalized by the standard deviation of each feature and then averaged across all features).
For models that can perform multiple imputation, 10 imputations are drawn for each test point to compute the average NRMSE scores.
For our model trained with the auxiliary objective (ACFlow+BG), we use the single ``best guess'' to compute the NRMSE. In order to not bias towards any specific missing pattern, we report the mean and standard deviations over 5 randomly generated binary masks (std. dev. are reported in Appendix Table.~\ref{tab:imputation}).

Quantitatively, \method is comparable to the previous state-of-the-art when trained purely by maximizing the likelihood. However, training with the auxiliary objective improves the NRMSE scores significantly and gives state-of-the-art results on all datasets considered.
As expected, higher missing rate makes it harder to learn the dependencies; however, our model performs best even when the missing rate is relatively high.

\section{Conclusion}

In this work, we demonstrated that we can model the arbitrary conditional distributions $p(x_u \mid x_o)$ using a single model by leveraging conditional flow transformations and conditional autoregressive likelihoods. As special cases, \method can also achieve good performance for modeling the joint distributions and arbitrary marginal distributions. 
In regard to applications, \method is applied to imputation tasks and it empirically outperforms several strong baselines. 
We also considered performing both single and multiple imputations in a unified platform to provide a ``best guess'' single imputation when the mean is not analytically available. The samples generated from our model show that we improve in both diversity and quality of imputations in many datasets. Our model typically recovers more details than the previous state-of-the-art methods.
In future work, we will apply \method to reason about causal relationships and learn underlying graphical structures. 
Our code is available at \url{https://github.com/lupalab/ACFlow}.

\section*{Acknowledgements}

This work was supported in part by the NIH 1R01AA026879-01A1 grant.

\bibliographystyle{icml2020}
\bibliography{main}

\begin{thebibliography}{33}
\providecommand{\natexlab}[1]{#1}
\providecommand{\url}[1]{\texttt{#1}}
\expandafter\ifx\csname urlstyle\endcsname\relax
  \providecommand{\doi}[1]{doi: #1}\else
  \providecommand{\doi}{doi: \begingroup \urlstyle{rm}\Url}\fi

\bibitem[Behrmann et~al.(2018)Behrmann, Duvenaud, and
  Jacobsen]{behrmann2018invertible}
Behrmann, J., Duvenaud, D., and Jacobsen, J.-H.
\newblock Invertible residual networks.
\newblock \emph{arXiv preprint arXiv:1811.00995}, 2018.

\bibitem[Belghazi et~al.(2019)Belghazi, Oquab, LeCun, and
  Lopez-Paz]{belghazi2019learning}
Belghazi, M.~I., Oquab, M., LeCun, Y., and Lopez-Paz, D.
\newblock Learning about an exponential amount of conditional distributions.
\newblock \emph{arXiv preprint arXiv:1902.08401}, 2019.

\bibitem[Butz et~al.(2019)Butz, Oliveira, dos Santos, and
  Teixeira]{butz2019deep}
Butz, C.~J., Oliveira, J.~S., dos Santos, A.~E., and Teixeira, A.~L.
\newblock Deep convolutional sum-product networks.
\newblock In \emph{Proceedings of the AAAI Conference on Artificial
  Intelligence}, volume~33, pp.\  3248--3255, 2019.

\bibitem[Buuren \& Groothuis-Oudshoorn(2010)Buuren and
  Groothuis-Oudshoorn]{micebuuren2010}
Buuren, S.~v. and Groothuis-Oudshoorn, K.
\newblock Mice: Multivariate imputation by chained equations in r.
\newblock In \emph{Journal of statistical software}, pp.\  1--68, 2010.

\bibitem[Chen et~al.(2016)Chen, Duan, Houthooft, Schulman, Sutskever, and
  Abbeel]{chen2016infogan}
Chen, X., Duan, Y., Houthooft, R., Schulman, J., Sutskever, I., and Abbeel, P.
\newblock Infogan: Interpretable representation learning by information
  maximizing generative adversarial nets.
\newblock In \emph{Advances in neural information processing systems}, pp.\
  2172--2180, 2016.

\bibitem[Dinh et~al.(2014)Dinh, Krueger, and Bengio]{dinh2014nice}
Dinh, L., Krueger, D., and Bengio, Y.
\newblock Nice: Non-linear independent components estimation.
\newblock \emph{arXiv preprint arXiv:1410.8516}, 2014.

\bibitem[Dinh et~al.(2016)Dinh, Sohl-Dickstein, and Bengio]{dinh2016density}
Dinh, L., Sohl-Dickstein, J., and Bengio, S.
\newblock Density estimation using real nvp.
\newblock \emph{arXiv preprint arXiv:1605.08803}, 2016.

\bibitem[Douglas et~al.(2017)Douglas, Zarov, Gourgoulias, Lucas, Hart, Baker,
  Sahani, Perov, and Johri]{douglas2017marginalizer}
Douglas, L., Zarov, I., Gourgoulias, K., Lucas, C., Hart, C., Baker, A.,
  Sahani, M., Perov, Y., and Johri, S.
\newblock A universal marginalizer for amortized inference in generative
  models.
\newblock \emph{arXiv preprint arXiv:1711.00695}, 2017.

\bibitem[Gondara \& Wang(2018)Gondara and Wang]{gondara2018mida}
Gondara, L. and Wang, K.
\newblock Mida: Multiple imputation using denoising autoencoders.
\newblock In \emph{Pacific-Asia Conference on Knowledge Discovery and Data
  Mining}, pp.\  260--272. Springer, 2018.

\bibitem[He et~al.(2016)He, Zhang, Ren, and Sun]{he2016deep}
He, K., Zhang, X., Ren, S., and Sun, J.
\newblock Deep residual learning for image recognition.
\newblock In \emph{Proceedings of the IEEE conference on computer vision and
  pattern recognition}, pp.\  770--778, 2016.

\bibitem[{Hore} \& {Ziou}(2010){Hore} and {Ziou}]{5596999}
{Hore}, A. and {Ziou}, D.
\newblock Image quality metrics: Psnr vs. ssim.
\newblock In \emph{2010 20th International Conference on Pattern Recognition},
  pp.\  2366--2369, Aug 2010.
\newblock \doi{10.1109/ICPR.2010.579}.

\bibitem[Houthooft et~al.(2016)Houthooft, Chen, Duan, Schulman, De~Turck, and
  Abbeel]{houthooft2016vime}
Houthooft, R., Chen, X., Duan, Y., Schulman, J., De~Turck, F., and Abbeel, P.
\newblock Vime: Variational information maximizing exploration.
\newblock In \emph{Advances in Neural Information Processing Systems}, pp.\
  1109--1117, 2016.

\bibitem[Ivanov et~al.(2018)Ivanov, Figurnov, and
  Vetrov]{ivanov2018variational}
Ivanov, O., Figurnov, M., and Vetrov, D.
\newblock Variational autoencoder with arbitrary conditioning.
\newblock \emph{arXiv preprint arXiv:1806.02382}, 2018.

\bibitem[Kingma \& Dhariwal(2018)Kingma and Dhariwal]{kingma2018glow}
Kingma, D.~P. and Dhariwal, P.
\newblock Glow: Generative flow with invertible 1x1 convolutions.
\newblock In \emph{Advances in Neural Information Processing Systems}, pp.\
  10215--10224, 2018.

\bibitem[Larochelle \& Murray(2011)Larochelle and Murray]{larochelle2011neural}
Larochelle, H. and Murray, I.
\newblock The neural autoregressive distribution estimator.
\newblock In \emph{Proceedings of the Fourteenth International Conference on
  Artificial Intelligence and Statistics}, pp.\  29--37, 2011.

\bibitem[Ledig et~al.(2017)Ledig, Theis, Husz{\'a}r, Caballero, Cunningham,
  Acosta, Aitken, Tejani, Totz, Wang, et~al.]{ledig2017photo}
Ledig, C., Theis, L., Husz{\'a}r, F., Caballero, J., Cunningham, A., Acosta,
  A., Aitken, A., Tejani, A., Totz, J., Wang, Z., et~al.
\newblock Photo-realistic single image super-resolution using a generative
  adversarial network.
\newblock In \emph{Proceedings of the IEEE conference on computer vision and
  pattern recognition}, pp.\  4681--4690, 2017.

\bibitem[Li et~al.(2019{\natexlab{a}})Li, Jiang, and Marlin]{li2019misgan}
Li, S. C.-X., Jiang, B., and Marlin, B.
\newblock Misgan: Learning from incomplete data with generative adversarial
  networks.
\newblock \emph{arXiv preprint arXiv:1902.09599}, 2019{\natexlab{a}}.

\bibitem[Li et~al.(2017)Li, Liu, Yang, and Yang]{li2017generative}
Li, Y., Liu, S., Yang, J., and Yang, M.-H.
\newblock Generative face completion.
\newblock In \emph{Proceedings of the IEEE Conference on Computer Vision and
  Pattern Recognition}, pp.\  3911--3919, 2017.

\bibitem[Li et~al.(2019{\natexlab{b}})Li, Gao, and Oliva]{li2019forest}
Li, Y., Gao, T., and Oliva, J.
\newblock A forest from the trees: Generation through neighborhoods.
\newblock \emph{arXiv preprint arXiv:1902.01435}, 2019{\natexlab{b}}.

\bibitem[Little \& Rubin(2019)Little and Rubin]{little2019statistical}
Little, R.~J. and Rubin, D.~B.
\newblock \emph{Statistical analysis with missing data}, volume 793.
\newblock John Wiley \& Sons, 2019.

\bibitem[Mattei \& Frellsen(2019)Mattei and Frellsen]{mattei2019miwae}
Mattei, P.-A. and Frellsen, J.
\newblock Miwae: Deep generative modelling and imputation of incomplete data
  sets.
\newblock In \emph{International Conference on Machine Learning}, pp.\
  4413--4423, 2019.

\bibitem[Molina et~al.(2019)Molina, Vergari, Stelzner, Peharz, Subramani,
  Di~Mauro, Poupart, and Kersting]{molina2019spflow}
Molina, A., Vergari, A., Stelzner, K., Peharz, R., Subramani, P., Di~Mauro, N.,
  Poupart, P., and Kersting, K.
\newblock Spflow: An easy and extensible library for deep probabilistic
  learning using sum-product networks.
\newblock \emph{arXiv preprint arXiv:1901.03704}, 2019.

\bibitem[Murray et~al.(2018)]{murray2018multiple}
Murray, J.~S. et~al.
\newblock Multiple imputation: A review of practical and theoretical findings.
\newblock \emph{Statistical Science}, 33\penalty0 (2):\penalty0 142--159, 2018.

\bibitem[Oliva et~al.(2018)Oliva, Dubey, Zaheer, P{\'o}czos, Salakhutdinov,
  Xing, and Schneider]{oliva2018transformation}
Oliva, J.~B., Dubey, A., Zaheer, M., P{\'o}czos, B., Salakhutdinov, R., Xing,
  E.~P., and Schneider, J.
\newblock Transformation autoregressive networks.
\newblock \emph{arXiv preprint arXiv:1801.09819}, 2018.

\bibitem[Papamakarios et~al.(2017)Papamakarios, Pavlakou, and
  Murray]{papamakarios2017masked}
Papamakarios, G., Pavlakou, T., and Murray, I.
\newblock Masked autoregressive flow for density estimation.
\newblock In \emph{Advances in Neural Information Processing Systems}, pp.\
  2338--2347, 2017.

\bibitem[Pathak et~al.(2016)Pathak, Krahenbuhl, Donahue, Darrell, and
  Efros]{pathak2016context}
Pathak, D., Krahenbuhl, P., Donahue, J., Darrell, T., and Efros, A.~A.
\newblock Context encoders: Feature learning by inpainting.
\newblock In \emph{Proceedings of the IEEE conference on computer vision and
  pattern recognition}, pp.\  2536--2544, 2016.

\bibitem[Poon \& Domingos(2011)Poon and Domingos]{poon2011sum}
Poon, H. and Domingos, P.
\newblock Sum-product networks: A new deep architecture.
\newblock In \emph{2011 IEEE International Conference on Computer Vision
  Workshops (ICCV Workshops)}, pp.\  689--690. IEEE, 2011.

\bibitem[Sajjadi et~al.(2018)Sajjadi, Bachem, Lucic, Bousquet, and
  Gelly]{sajjadi2018assessing}
Sajjadi, M.~S., Bachem, O., Lucic, M., Bousquet, O., and Gelly, S.
\newblock Assessing generative models via precision and recall.
\newblock In \emph{Advances in Neural Information Processing Systems}, pp.\
  5228--5237, 2018.

\bibitem[Salimans et~al.(2017)Salimans, Karpathy, Chen, and
  Kingma]{salimans2017pixelcnn++}
Salimans, T., Karpathy, A., Chen, X., and Kingma, D.~P.
\newblock Pixelcnn++: Improving the pixelcnn with discretized logistic mixture
  likelihood and other modifications.
\newblock \emph{arXiv preprint arXiv:1701.05517}, 2017.

\bibitem[Stekhoven \& B{\"u}hlmann(2011)Stekhoven and
  B{\"u}hlmann]{stekhoven2011missforest}
Stekhoven, D.~J. and B{\"u}hlmann, P.
\newblock Missforest—non-parametric missing value imputation for mixed-type
  data.
\newblock \emph{Bioinformatics}, 28\penalty0 (1):\penalty0 112--118, 2011.

\bibitem[Stekhoven \& Bühlmann(2012)Stekhoven and
  Bühlmann]{missstekhoven2012}
Stekhoven, D.~J. and Bühlmann, P.
\newblock Missforest — non-parametric missing value imputation for mixed-type
  data.
\newblock In \emph{Bioinformatics, Volume 28, Issue 1}, pp.\  112--118, 2012.

\bibitem[Troyanskaya et~al.(2001)Troyanskaya, Cantor, Sherlock, Brown, Hastie,
  Tibshirani, Botstein, and Altman]{troyanskaya2001missing}
Troyanskaya, O., Cantor, M., Sherlock, G., Brown, P., Hastie, T., Tibshirani,
  R., Botstein, D., and Altman, R.~B.
\newblock Missing value estimation methods for dna microarrays.
\newblock \emph{Bioinformatics}, 17\penalty0 (6):\penalty0 520--525, 2001.

\bibitem[Yoon et~al.(2016)Yoon, Jordon, and Schaar]{yoon2018gain}
Yoon, J., Jordon, J., and Schaar, M. v.~d.
\newblock Gain: Missing data imputation using generative adversarial nets.
\newblock \emph{arXiv preprint arXiv:1605.08803}, 2016.

\end{thebibliography}

\clearpage
\newpage
\setcounter{figure}{0}
\setcounter{table}{0}
\setcounter{footnote}{0}
\renewcommand\thefigure{\thesection.\arabic{figure}}
\renewcommand\thetable{\thesection.\arabic{table}}

\appendix
\appendixpage

\section{\method for Categorical Data}\label{sec:categorical}
In the main text, we mainly focus on real-valued data, but our model is also applicable with data that contains categorical dimensions. Consider a $d$-dimensional vector $x$ with both real-valued and categorical features. The conditional distribution $p(x_u | x_o)$ can be factorized as $p(x_u^c|x_o)p(x_u^r|x_o,x_u^c)$, where $x_u^r$ and $x_u^c$ represent the real-valued and categorical components in $x_u$ respectively. Since the conditioning inputs in Eq.~\eqref{eq:condtrans} can be either real-valued or categorical, $p(x_u^r|x_o,x_u^c)$ can be directly modeled by \method. 
However, \method is not directly applicable to $p(x_u^c|x_o)$, because the change of variable theorem, Eq. \eqref{eq:changevariables}, cannot be applied to categorical covariates. But we can use an autoregressive model, i.e., a special \method without any transformations, to get likelihoods for categorical components. Hence, our proposed method can be applied to both real-valued and categorical data.

\section{Synthetic Datasets}\label{sec:syn}

To validate the effectiveness of our model, we conduct experiments on synthetic 2-dimensional datasets \cite{behrmann2018invertible}, i.e., $x=(x_1,x_2) \in \mathbb{R}^2$. The joint distributions are plotted in Fig.~\ref{fig:syn}.
Here, the conditional distributions are highly multi-modal and vary significantly when conditioned on different observations. We train arbitrary conditional models on 100,000 samples from the joint distribution. The masks are generated by dropping out one dimension at random.

We compare our model against VAEAC, the current \emph{state-of-the-art} model trained purely by likelihood. 
We closely follow the released code for VAEAC\footnote{\url{https://github.com/tigvarts/vaeac/blob/master/imputation_networks.py}} to construct the proposal, prior, and generative networks. Specifically, we use fully connected layers and skip connections as is in their official implementation. We also use short-cut connections between the prior network and the generative network. We search over different combinations of the number of layers, the number of hidden units of fully connected layers, and the dimension of the latent code. Validation likelihoods are used to select the best model.

\begin{figure}[htb]
    \centering
    \includegraphics[width=0.5\linewidth]{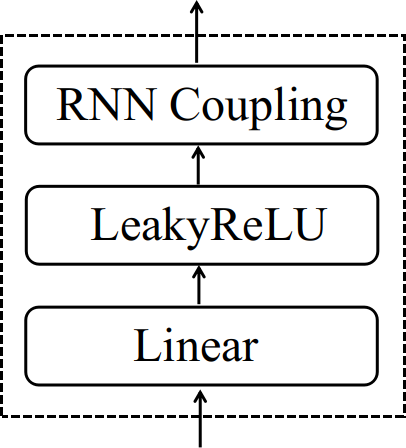}
    \caption{One conditional transformation layer.}
    \label{fig:layer}
\end{figure}

\method is constructed by stacking multiple conditional transformations and an autoregressive likelihood. One conditional transformation layer (shown in Fig.~\ref{fig:layer}) contains one linear transformation, followed by one leaky relu transformation, and then one RNN coupling transformation. The DNN in the linear transformation is a 2-layer fully connected network with 256 units. RNN coupling is implemented as a 2-layer GRU with 256 hidden units. We stack four conditional transformation layers and use reverse transformations in between. The autoregressive likelihood model is a 2-layer GRU with 256 hidden units. The base distribution is a Gaussian Mixture Model with 40 components.

Fig.~\ref{fig:syn} shows imputation results from our model and VAEAC. We show imputations from both $p(x_1 \mid x_2, b)$ and $p(x_2 \mid x_1,b)$ by generating 10 samples conditioned on each observed covariate. Observed covariates are sampled from the marginal distribution $p(x_2)$ and $p(x_1)$ respectively.
We see that \method is capable of learning multi-modal distributions, while VAEAC tends to merge multiple modes into one single mode.

\section{Image Datasets}\label{sec:inpaint}

\subsection{Datasets and Masks}\label{sec:inpaint_data}
We compare \method against baselines using three common image datasets, MNIST, Omniglot, and CelebA. MNIST contains grayscale images of size $28 \times 28$. Omniglot data is also resized to $28 \times 28$ and augmented with rotations. For CelebA dataset, we take a central crop of $128 \times 128$ and then resize it to $64 \times 64$. Since a flow model requires continuous inputs, we dequantize the images by adding independent uniform noise to them.

In order to train the conditional model $p(x_u \mid x_0, b)$, we manually generate the binary masks form some predefined distributions. We train \method and baselines using a mixture of different masks. For MNIST and Omniglot, we use MCAR masks, rectangular masks, and half masks. For CelebA, other than the three masks described above, we also use a random pattern mask proposed in \cite{pathak2016context} and the masks used in \cite{li2017generative}. We shall describe the meaning of different masks below.

\paragraph{MCAR mask}
MCAR mask utilizes a pixelwise independent Bernoulli distribution with $p=0.2$ to construct masks. That is, on average, we randomly drop out 80\% of the pixels. 

\paragraph{Rectangular mask}
We randomly generate a rectangle inside which pixels are marked as unobserved. The mask is generated by sampling a point to be the upper-left corner and randomly generating the height and width of the rectangle, although the area is required to be at least 10\% of the full image area.

\paragraph{Half mask}
Half mask means either the upper, lower, left or right half of the image is unobserved.

\paragraph{Random pattern mask}
Random pattern means we mask out a random region with an arbitrary shape. We take the implementation from VAEAC\footnote{\url{https://github.com/tigvarts/vaeac/blob/master/mask_generators.py##L100}}.

\paragraph{Masks in \cite{li2017generative}}
They proposed six different masks to mask out different parts of a face, like eyes, mouth and checks.

\subsection{Models}\label{sec:inpaint_model}

In this experiment, we compare to VAEAC \cite{ivanov2018variational} and an autoregressive model, PixelCNN \cite{salimans2017pixelcnn++}.

VAEAC released their implementation for the $128 \times 128$ CelebA dataset and we modify their code by removing the first pooling layer to suit $64 \times 64$ images. For MNIST and Omniglot, we build similar networks by using fewer convolution and pooling layers. Specifically, we pad the images to $32 \times 32$ and use 5 pooling layers to get a latent vector. We use 32 latent variables for MNIST and Omniglot and 256 latent variables for CelebA. We also use short-cut connections between the prior network and the generative network as in \cite{ivanov2018variational}.

PixelCNN is originally proposed to model the joint likelihoods. However, since it decomposes the joint likelihood into a sequence of conditionals, it can be easily extended to model the arbitrary conditional distributions:
\begin{equation}
    p(x_u \mid x_o, b) = \prod_{i=1}^{|u|} p(x_u^{i} \mid x_u^{<i}, x_o, b),
\end{equation}
where $x_u^i$ represents the $ith$ dimension of $x_u$. In order to capture the dependencies over $x_o$ and $b$, we apply an embedding network to get a vector embedding of them. Note that different from the original PixelCNN implementation, where they use discrete base distributions, we use mixture of Gaussian base distribution here to make the comparison fair. 

\method is implemented by replacing affine coupling in RealNVP \cite{dinh2016density} with our proposed conditional affine coupling transformation. The DNN in affine coupling is implemented as a ResNet \cite{he2016deep}. For MNIST and Omniglot, we use 2 residual blocks. For CelebA, we use 4 blocks. We also use the multi scale architecture via the squeeze operation proposed in \cite{dinh2016density}. For MNIST and Omniglot, we apply 3 squeeze operations. For CelebA, we apply 4 squeeze operations. Note that we also need to squeeze the binary mask $b$ in the same way to make sure it corresponds to $x_o$. 
The arbitrary conditional likelihood is chosen as either a Gaussian base likelihood model or an autoregressive one with the same formulation as the PixelCNN model described above.

For models trained with our proposed ``best guess'' penalty, we set the hyperparameter $\lambda$ to 1.0. During preliminary experiments, we find our model is quite robust to different $\lambda$ values.

\subsection{Joint Likelihood}\label{sec:joint}

\begin{figure*}[htb]
    \centering
    \subfigure[MNIST]{
    \includegraphics[width=0.32\linewidth]{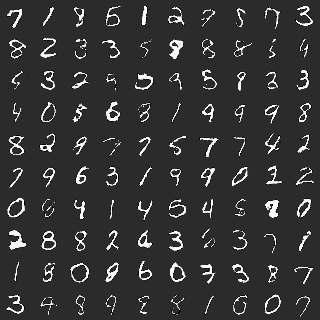}}
    \subfigure[Omniglot]{
    \includegraphics[width=0.32\linewidth]{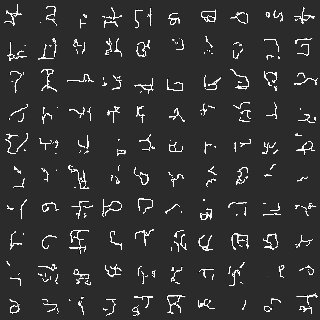}}
    \subfigure[CelebA]{
    \includegraphics[width=0.32\linewidth]{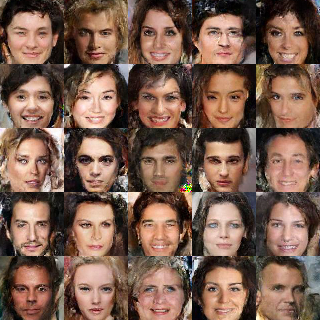}}
    \caption{Joint sampling from a model trained with $p(x_u \mid x_o, b)$ by setting $b$ to zeros.}
    \label{fig:inpaint_joint}
\end{figure*}

For a model trained with arbitrary conditional likelihood $p(x_u \mid x_o, b)$, we evaluate the joint likelihood by setting the binary mask $b$ to zeros. We can of course sample from this joint distribution as well. We show some samples in Fig.~\ref{fig:inpaint_joint}. We see that the model generate decent samples although it never observes the complete data during training.

\subsection{Gibbs Sampling}\label{sec:gibbs}

\begin{figure*}
    \centering
    \includegraphics[width=\linewidth]{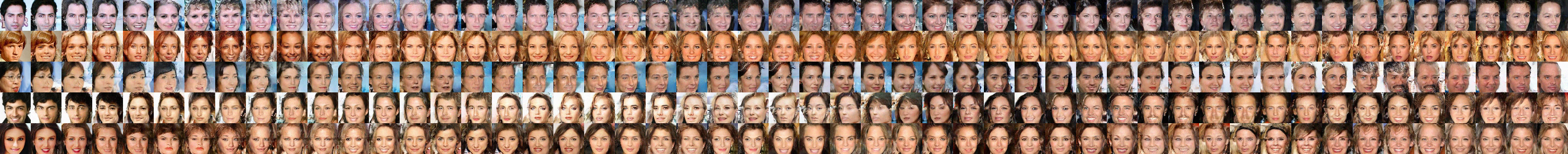}
    \caption{Gibbs sampling using a pretrained ACFlow. We iteratively sample the upper and lower half for 50 steps conditioned on the remaining part.}
    \label{fig:gibbs_50}
\end{figure*}

Due to space limitation, we only show part of the mixing steps in the Gibbs sampling procedure. In Figure.~\ref{fig:gibbs_50}, we show all the 50 steps. Please zoom in for better visualization.

\subsection{Additional Inpaintings}\label{sec:inpaint_sample}
\begin{figure*}[htb]
    \centering
    \subfigure[MNIST]{
    \includegraphics[height=0.3\linewidth]{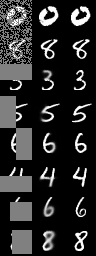}
    }
    \subfigure[Omniglot]{
    \includegraphics[height=0.3\linewidth]{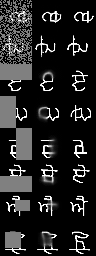}
    }
    \subfigure[CelebA]{
    \includegraphics[height=0.3\linewidth]{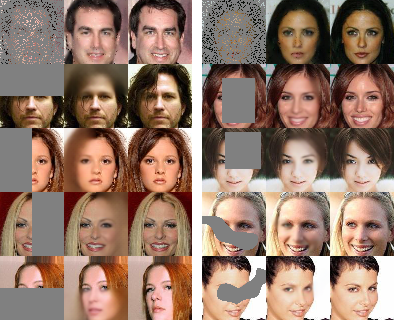}
    }
    \caption{Single imputation from our ``best guess''. Left: inputs. Middle: best guess imputation. Right: groundtruth.}
    \label{fig:single}
\end{figure*}

\begin{figure*}[htb]
    \centering
    \subfigure[MNIST]{
    \includegraphics[width=0.4\linewidth]{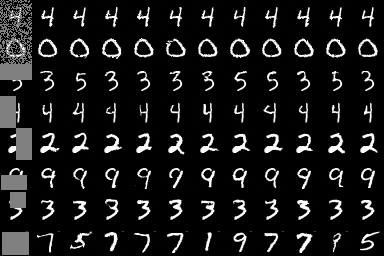}
    }
    \subfigure[Omniglot]{
    \includegraphics[width=0.4\linewidth]{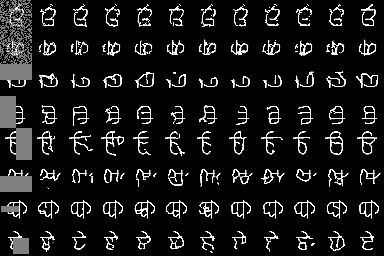}
    }
    \subfigure[CelebA]{
    \includegraphics[width=0.8\linewidth]{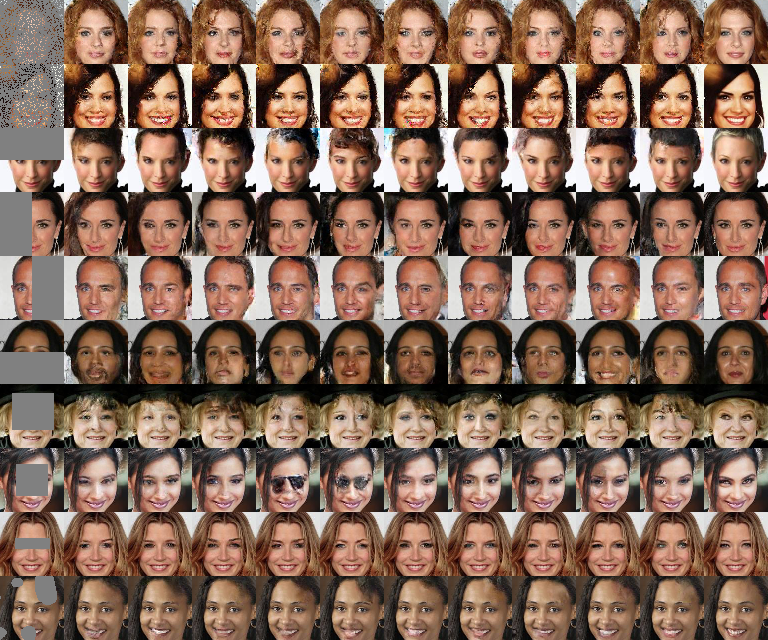}
    }
    \caption{Additional inpaintings from \method. Left: inputs. Middle: Samples. Right: groundtruth.}
    \label{fig:additional}
\end{figure*}

\begin{figure*}
    \centering
    \subfigure[MNIST]{
    \includegraphics[width=0.45\linewidth]{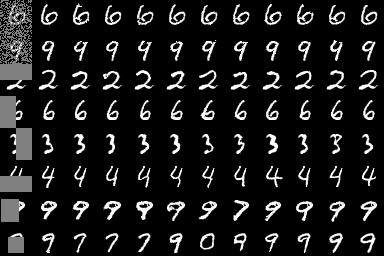}
    }
    \subfigure[Omniglot]{
    \includegraphics[width=0.45\linewidth]{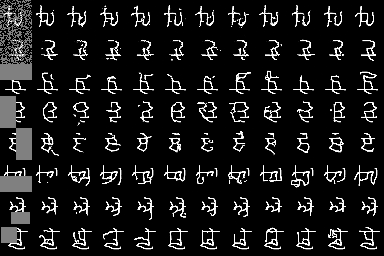}
    }
    \subfigure[CelebA]{
    \includegraphics[width=0.9\linewidth]{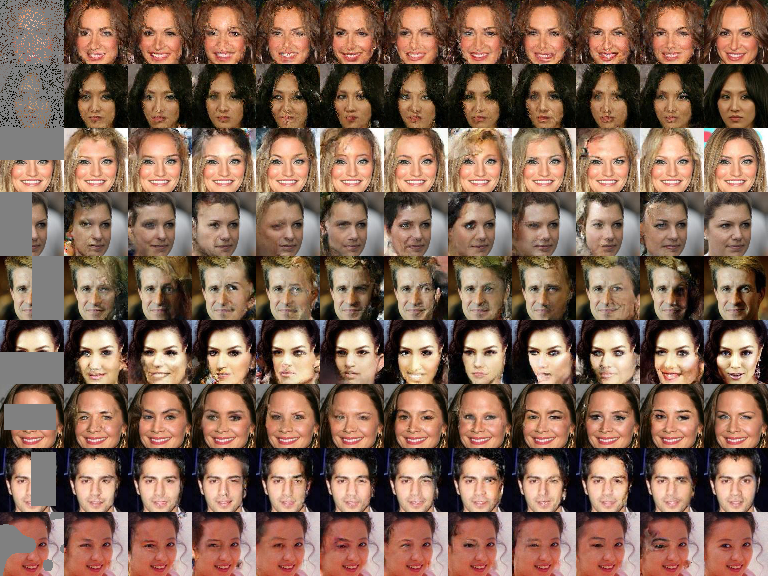}
    }
    \caption{Multiple imputations from ACFlow+BG. Left:inputs. Middle: Samples. Right: groundtruth.}
    \label{fig:lambda}
\end{figure*}

We show additional inpainting results from \method in Fig.~\ref{fig:additional}. We also show some ``best guess'' inpaintings obtained by a model trained with the auxiliary objective in Fig.~\ref{fig:single}. As we expected, the ``best guess'' imputations tend to be blurry due to the MSE penalty. However, the samples from the same model are still diverse and coherent as can be seen from Fig.~\ref{fig:lambda}.

\section{Real-valued Datasets}\label{sec:imputation}

\subsection{Models}\label{sec:imputation_model}
In this experiment, we compare to VAEAC and an autoregressive model in term of the likelihood. We also compare to a GAN based method, GAIN, and two classic imputation methods, MICE and MisForest in terms of the imputation performance.

We modify the official VAEAC implementation for this experiment. In their original implementation, they restricted the generative network to output a distribution with variance equal to $\mathbf{1}$. We found that learning the variance can improve VAEAC's likelihood significantly and gives comparable NRMSE scores. Note that during sampling, we still use the mean of the outputs from the generative network, because it gives slightly better NRMSE scores.

The autoregressive baseline is implemented as a 4-layer GRU netowrk with 256 hidden units. We did not observe further improvements when adding more layers.

We build \method by combining 6 conditional transformation layers (shown in Fig.~\ref{fig:layer}) with a 4-layer GRU autoregerssive likelihood model. The base distribution is a Gaussian Mixture with 40 components. ``ACFlow+BG'' uses the same network architecture but trained with $\lambda=1.0$. We search over 0.1, 1.0, and 10.0 for $\lambda$ and find it gives comparable likelihood for all datasets. Hence, we only report results when $\lambda=1.0$ in the main text.

For GAIN, we use the variant the VAEAC authors proposed. They observed consistent improvement over the original one on UCI datasets. When we have complete training data, we add another MSE loss over the unobserved covariates to train the generator.

MICE and MissForest are trained using default parameters in the R packages.

The autoencoder baseline is implemented as a 6-layer fully connected network with 256 hidden units. ReLU is used after each fully connected layer except the last one. We use standard Gaussian as the base distribution.

\subsection{Marginal Likelihood}

\begin{figure*}
    \centering
    \subfigure[bsds]{
    \includegraphics[width=0.3\linewidth,trim=3cm 1cm 1.5cm 1.5cm,clip]{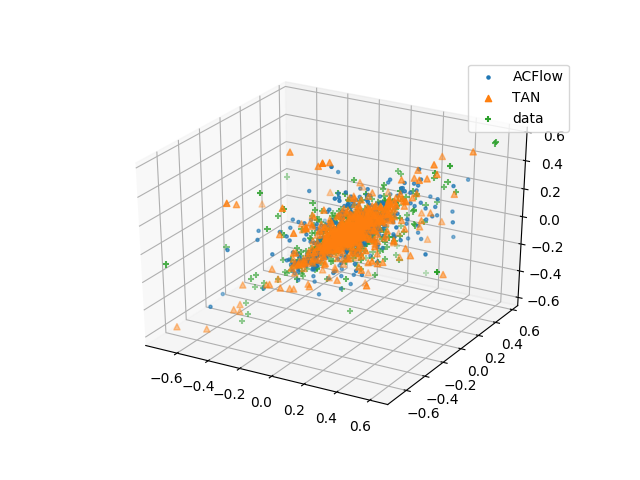}}
    \subfigure[gas]{
    \includegraphics[width=0.3\linewidth,trim=3cm 1cm 1.5cm 1.5cm,clip]{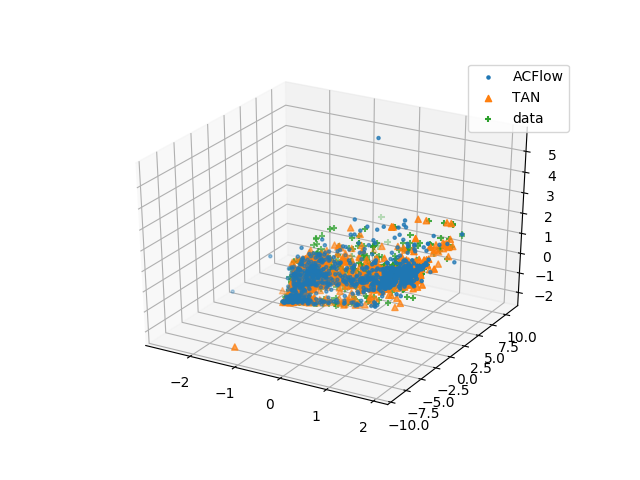}}
    \subfigure[hepmass]{
    \includegraphics[width=0.3\linewidth,trim=3cm 1cm 1.5cm 1.5cm,clip]{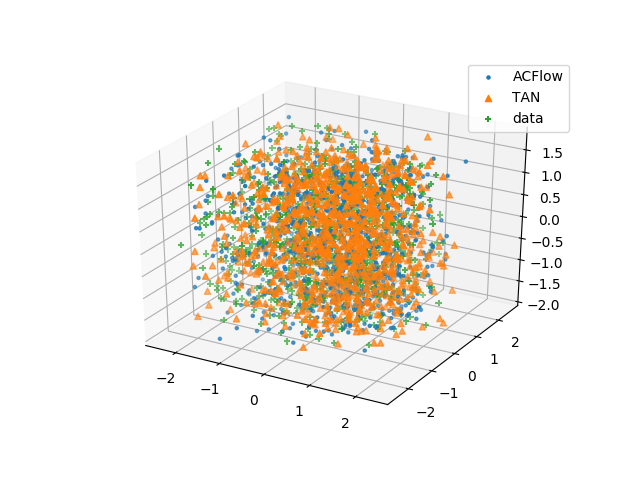}}
    \subfigure[miniboone]{
    \includegraphics[width=0.3\linewidth,trim=3cm 1cm 1.5cm 1.5cm,clip]{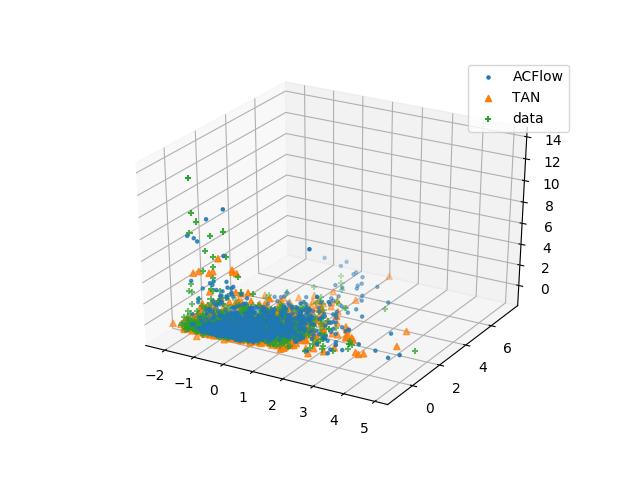}}
    \subfigure[power]{
    \includegraphics[width=0.3\linewidth,trim=3cm 1cm 1.5cm 1.5cm,clip]{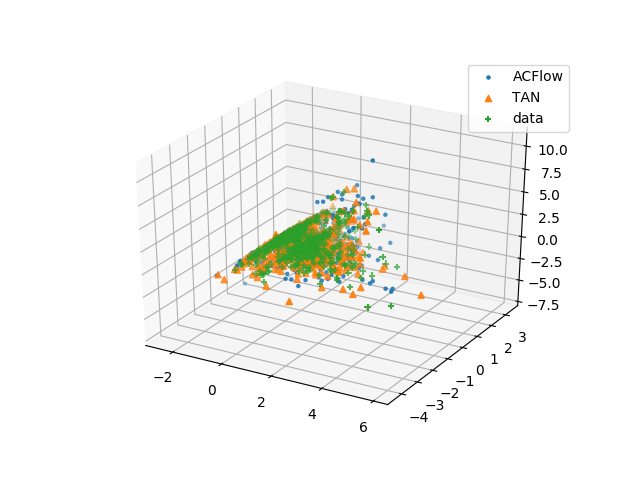}}
    \caption{sample the first 3 dimensions from the learned marginal distributions.}
    \label{fig:margin}
\end{figure*}

In the main text, we compare the marginal likelihood to TANs that are trained specifically for the corresponding marginal distributions. In Figure.~\ref{fig:margin}, we qualitatively compare them by showing the scatter plots of the samples. We draw 1024 samples from both ACFlow and TANs, and also sample 1024 real data.

\subsection{Imputation Results}
We list all imputation results in Table.~\ref{tab:imputation}. We report mean and standard deviation by generating 5 different masks for each test data point.

\begin{table*}
    \caption{Missing feature imputation results. Lower is better for both NRMSE and NLL. Numbers inside a parentheses are standard deviation for 5 randomly generated binary mask.}
    \label{tab:imputation}
    \centering
    \small
    \begin{tabular}{c|cc|ccccc}
    \toprule
    Missing Rate & Method & & bsds & gas & hepmass & miniboone & power \\
    \midrule
    \multirow{11}{*}{p=0} & GAIN & NRMSE & 0.895\tabpm0.151 & 0.715\tabpm0.041 & 0.948\tabpm0.006 & 0.620\tabpm0.002 & 0.949\tabpm0.017 \\ \cmidrule{2-3}
                         & \multirow{2}{*}{AutoEncoder} & NRMSE & 0.635\tabpm0.000 & 1.016\tabpm0.178 & 0.930\tabpm0.001 & 0.483\tabpm0.003 & 0.887\tabpm0.002 \\ \cmidrule{3-3}
                         &                              & NLL & 3.502\tabpm0.014 & -2.543\tabpm0.115 & 12.707\tabpm0.018 & 6.861\tabpm0.114 & 2.488\tabpm0.039 \\  \cmidrule{2-3}
                         & \multirow{2}{*}{VAEAC} & NRMSE & 0.615\tabpm0.000 & 0.574\tabpm0.033 & 0.896\tabpm0.001 & 0.462\tabpm0.002 & 0.880\tabpm0.001 \\ \cmidrule{3-3}
                         &                        & NLL & 1.708\tabpm0.005 & -2.418\tabpm0.006 & 10.082\tabpm0.010 & 3.452\tabpm0.067 & 0.042\tabpm0.002 \\ \cmidrule{2-3}
                         & \multirow{2}{*}{AutoRegressive} & NRMSE & 0.652\tabpm0.000 & 0.457\tabpm0.011 & 0.903\tabpm0.001 & 0.460\tabpm0.005 & 0.877\tabpm0.002\\ \cmidrule{3-3}
                         &                          & NLL & -1.73\tabpm0.010 & -7.646\tabpm0.009 & 6.428\tabpm0.006 & -0.057\tabpm0.022 & -0.399\tabpm0.003\\  \cmidrule{2-3}
                         & \multirow{2}{*}{ACFlow} & NRMSE & 0.603\tabpm0.000 & 0.567\tabpm0.050 & 0.909\tabpm0.000  & 0.478\tabpm0.004 & 0.877\tabpm0.001 \\ \cmidrule{3-3}
                         &                                  & NLL & -5.269\tabpm0.007 & -8.086\tabpm0.010 & 8.197\tabpm0.008 & 0.972\tabpm0.022 & -0.561\tabpm0.003\\ \cmidrule{2-3}
                         & \multirow{2}{*}{ACFlow+BG} & NRMSE & 0.572\tabpm0.000 & 0.369\tabpm0.016 & 0.861\tabpm0.001 & 0.442\tabpm0.001 & 0.833\tabpm0.002\\ \cmidrule{3-3}
                         &                                  & NLL & -4.841\tabpm0.008 & -7.593\tabpm0.011 & 6.833\tabpm0.006 & 1.098\tabpm0.032 & -0.528\tabpm0.003\\
    \midrule
    \multirow{13}{*}{p=0.1} & MICE       & NRMSE &0.631\tabpm0.003 & 0.518\tabpm0.004 & 0.964\tabpm0.004 & 0.605\tabpm0.004 & 0.911\tabpm0.008 \\ \cmidrule{2-3}
                           & MissForest & NRMSE &0.665\tabpm0.002 & 0.418\tabpm0.005 & 0.985\tabpm0.002 & 0.561\tabpm0.003 & 0.991\tabpm0.019\\ \cmidrule{2-3}
                           & GAIN       & NRMSE & 0.749\tabpm0.128 & 0.502\tabpm0.070 & 1.024\tabpm0.023 & 0.615\tabpm0.017 & 1.074\tabpm0.038 \\ \cmidrule{2-3}
                           & \multirow{2}{*}{AutoEncoder} & NRMSE & 0.648\tabpm0.001 & 0.761\tabpm0.095 & 0.936\tabpm0.001 & 0.498\tabpm0.002 & 0.887\tabpm0.002 \\ \cmidrule{3-3}
                           &                              & NLL & 4.300\tabpm0.038 & -2.266\tabpm0.169 & 12.851\tabpm0.012 & 7.305\tabpm0.043 & 2.521\tabpm0.028 \\ \cmidrule{2-3}
                           & \multirow{2}{*}{VAEAC} & NRMSE & 0.620\tabpm0.000 & 0.558\tabpm0.047 & 0.899\tabpm0.000 & 0.467\tabpm0.004 & 0.881\tabpm0.003 \\ \cmidrule{3-3}
                           &                        & NLL & 2.245\tabpm0.015 & -2.823\tabpm0.009 & 10.389\tabpm0.005 & 4.242\tabpm0.071 & 0.103\tabpm0.005 \\ \cmidrule{2-3}
                           & \multirow{2}{*}{AutoRegressive} & NRMSE & 0.752\tabpm0.000 & 0.472\tabpm0.011 & 0.915\tabpm0.000 & 0.539\tabpm0.004 & 0.876\tabpm0.001\\ \cmidrule{3-3}
                           &                          & NLL & 3.233\tabpm0.015 & -7.536\tabpm0.009 & 7.824\tabpm0.006 & 5.409\tabpm0.066 & -0.466\tabpm0.003\\  \cmidrule{2-3}
                           & \multirow{2}{*}{ACFlow} & NRMSE & 0.610\tabpm0.000 & 0.588\tabpm0.025 & 0.908\tabpm0.001 & 0.533\tabpm0.005 & 0.877\tabpm0.002 \\ \cmidrule{3-3}
                           &                                  & NLL & -4.225\tabpm0.018 & -7.568\tabpm0.005 & 7.784\tabpm0.006 & 5.150\tabpm0.053 & -0.557\tabpm0.003 \\ \cmidrule{2-3}
                           & \multirow{2}{*}{ACFlow+BG} & NRMSE & 0.586\tabpm0.001 & 0.384\tabpm0.018 & 0.863\tabpm0.001 & 0.468\tabpm0.003 & 0.836\tabpm0.002 \\ \cmidrule{3-3}
                           &                                  & NLL & -3.187\tabpm0.017 & -7.212\tabpm0.008 & 9.670\tabpm0.007 & 3.577\tabpm0.057 & -0.510\tabpm0.003 \\
    \midrule
    \multirow{13}{*}{p=0.5} & MICE       & NRMSE & 0.628\tabpm0.001 & 0.539\tabpm0.005  & 0.969\tabpm0.002 & 0.615\tabpm0.002 & 0.916\tabpm0.005\\ \cmidrule{2-3}
                           & MissForest & NRMSE & 0.662\tabpm0.001 & 0.436\tabpm0.003 & 0.990\tabpm0.002 & 0.573\tabpm0.005 & 0.990\tabpm0.012 \\ \cmidrule{2-3}
                           & GAIN       & NRMSE & 0.929\tabpm0.123 & 1.152\tabpm0.180 & 1.143\tabpm0.035 & 0.800\tabpm0.042 & 1.101\tabpm0.044 \\ \cmidrule{2-3}
                           & \multirow{2}{*}{AutoEncoder} & NRMSE & 0.739\tabpm0.001 & 0.618\tabpm0.056 & 0.962\tabpm0.001 & 0.567\tabpm0.005 & 0.905\tabpm0.002 \\ \cmidrule{3-3}
                           &                              & NLL & 10.078\tabpm0.021 & 0.990\tabpm0.097 & 13.482\tabpm0.012 & 10.775\tabpm0.091 & 2.858\tabpm0.047 \\ \cmidrule{2-3}
                           & \multirow{2}{*}{VAEAC} & NRMSE & 0.666\tabpm0.001 & 0.531\tabpm0.036 & 0.915\tabpm0.001 & 0.513\tabpm0.004 & 0.892\tabpm0.002 \\ \cmidrule{3-3}
                           &                        & NLL & 9.930\tabpm0.029 & -1.952\tabpm0.023 & 11.415\tabpm0.012 & 9.051\tabpm0.079 & 0.343\tabpm0.004\\ \cmidrule{2-3}
                           & \multirow{2}{*}{AutoRegressive} & NRMSE & 0.879\tabpm0.001 & 0.483\tabpm0.021& 0.937\tabpm0.000 & 0.644\tabpm0.002 & 0.882\tabpm0.002\\ \cmidrule{3-3}
                           &                          & NLL & 11.348\tabpm0.010 & -5.723\tabpm0.004 & 9.760\tabpm0.007 & 11.024\tabpm0.069 & -0.363\tabpm0.003\\  \cmidrule{2-3}
                           & \multirow{2}{*}{ACFlow} & NRMSE & 0.667\tabpm0.001 & 0.488\tabpm0.030 & 0.938\tabpm0.000 & 0.614\tabpm0.004 & 0.890\tabpm0.000 \\ \cmidrule{3-3}
                           &                                  & NLL & 1.508\tabpm0.010 & -5.405\tabpm0.008 & 10.538\tabpm0.006 & 9.892\tabpm0.084 & -0.458\tabpm0.005 \\ \cmidrule{2-3}
                           & \multirow{2}{*}{ACFlow+BG} & NRMSE & 0.645\tabpm0.000 & 0.421\tabpm0.016 & 0.890\tabpm0.000 & 0.582\tabpm0.007 &0.843\tabpm0.001 \\ \cmidrule{3-3}
                           &                                  & NLL & 3.497\tabpm0.015 & -4.818\tabpm0.009 & 10.975\tabpm0.006 & 10.849\tabpm0.105 & -0.417\tabpm0.005 \\
    \bottomrule
    \end{tabular}
\end{table*}

\end{document}